\documentclass{article}

\usepackage[final,main,nonatbib]{neurips_2026}
\renewcommand{\acksection}{\section*{Acknowledgments}}
\usepackage[utf8]{inputenc}
\usepackage[T1]{fontenc}
\usepackage{lmodern}
\usepackage{microtype}
\usepackage{amsmath, amssymb, amsfonts}
\usepackage{nicefrac}
\usepackage{xcolor}
\usepackage{booktabs}
\usepackage{graphicx}
\usepackage{hyperref}
\usepackage{url}
\usepackage{xurl}
\usepackage{footnote}
\usepackage{float}
\usepackage{textcomp}
\usepackage{tikz}

\usepackage[numbers,sort&compress]{natbib}

\hypersetup{
    colorlinks=true,
    linkcolor=blue!50!black,
    citecolor=blue!50!black,
    urlcolor=blue!50!black
}


\title{Emergent Biological Capabilities in a Foundation Model \\ for Molecular Interactions}

\author{%
    Adi Mashiach \\
    Output Labs \\
    \texttt{adi@outputlabs.ai}
}

\begin{document}

\maketitle

\begin{abstract}
Foundation models trained on large unlabeled corpora develop emergent capabilities, tasks that go beyond the training objective. We extend this paradigm to multimodal biology with \textsc{Bonbon}, a foundation model for \emph{molecular interactions} trained on approximately 3.2 trillion tokens of paired protein-ligand sequences. From a single representation space, \textsc{Bonbon} exhibits three zero-shot emergent capabilities at hierarchical levels of resolution: \emph{functional} (mechanism of action), \emph{residue} (binding site localization, including orthosteric/allosteric discrimination), and \emph{bond} (covalent versus non-covalent engagement). All three capabilities generalize to out-of-distribution complexes. \textsc{Bonbon} was trained on protein-ligand sequences alone, without any task-specific supervision and without structural coordinates. We conducted controlled scaling experiments across six model scales from 42M to 1.6B parameters; every capability evaluated shows a sharp jump at the largest scale on continuous metrics, even as pretraining loss decreases smoothly. This demonstrates that loss-based scaling laws alone are insufficient to predict when zero-shot emergent capabilities become available. The capabilities reported here emerge after training at a 1:2{,}000 parameter-to-token ratio to accommodate the sparsity of paired biological interaction data. The same frozen representations achieve state-of-the-art binding and affinity prediction at five orders of magnitude greater speed than co-folding methods. In prospective wet-lab validation across nine disease targets spanning seven protein families, \textsc{Bonbon} achieves a 61.5\% dose-response hit rate on de novo compounds, yielding structurally novel molecules not previously reported to inhibit these proteins, including for out-of-distribution targets.
\end{abstract}

\section{Introduction}
\label{sec:intro}

Foundation models trained by self-supervised learning at scale exhibit emergent capabilities: tasks that go beyond the training objective~\citep{wei_2022b_emergent}. This principle extends from language~\citep{brown_2020_language, wei_2022a_chain} to multimodal vision-language~\citep{georgiev_2024_gemini, driess_2023_palm} and unimodal biological domains: ESM-2 predicts protein structure from masked language modeling~\citep{lin_2023_evolutionary}, and Evo captures regulatory patterns across genomes~\citep{nguyen_2024_sequence}. Yet no model for molecular interactions has demonstrated capabilities beyond its explicit training objective without task-specific supervision. Co-folding models~\citep{abramson_2024_alphafold3, passaro_2025_boltz, chai_2024_chai} have transformed protein-ligand structure prediction, yet their capabilities do not extend beyond predicting 3D atomic coordinates. Independent benchmarks report substantial memorization with limited generalization to unseen ligands~\citep{xu_2025_benchmarking, skrinjar_2025_have}; capabilities such as affinity prediction require supervised heads rather than emerging from the pretrained representation. The broader landscape of molecular interaction modeling remains fragmented, with task-specific models each trained with explicit supervision for a single capability~\citep{leguilloux_2009_fpocket, krivak_2018_p2rank, zhu_2014_docking}.

We hypothesize that a foundation model for molecular interactions, with an architecture matched to the learning objective and trained at sufficient scale, can extract high-order interaction properties encoded in the sequences of interacting molecules, without requiring structural coordinates or physics-based analysis. Molecular interaction modeling faces challenges in both data and architecture that distinguish it from vision-language foundation models. For example, paired protein-ligand data is sparse, carrying far less signal than a dense image-caption pair~\citep{peon_2017_predicting, yang_2019_analyzing}, with the patterns distinguishing functional outcomes (agonism vs.\ antagonism, covalent vs.\ non-covalent engagement) distributed across millions of examples rather than concentrated within each. Similarly, molecular interactions form many-to-many relationships, where one protein binds thousands of ligands and one ligand hundreds of proteins, whereas standard multimodal architectures assume a one-to-one correspondence between modalities.

We introduce \textsc{Bonbon}, a transformer-based model trained by self-supervised learning on approximately 3.2 trillion tokens of paired protein-ligand sequences. \textsc{Bonbon} adopts a data-heavy scaling regime: a 1:2{,}000 parameter-to-token ratio, two orders of magnitude beyond the Chinchilla-optimal ratio for natural language~\citep{hoffmann_2022_training}.

From a single representation space, \textsc{Bonbon} exhibits three emergent capabilities arising at hierarchical levels of resolution. At the \textbf{functional} level, \textsc{Bonbon} zero-shot predicts mechanism of action across six classes, including agonist/antagonist discrimination on the same receptor. At the \textbf{residue} level, \textsc{Bonbon} localizes where interactions occur, enabling zero-shot binding site detection, including on targets such as GPCRs and transporters that lack well-defined pockets, and a parameter-free signature distinguishes orthosteric from allosteric pockets, including on the same protein. At the \textbf{bond} level, dense representations classify covalent versus non-covalent engagement and per-residue representations localize the reactive residue. All three capabilities emerge from paired sequences alone, without task-specific labels, structural coordinates, or binding geometries, and generalize to out-of-distribution complexes. In a controlled scaling experiment across six model scales (42M to 1.6B parameters), all three capabilities show a sharp jump at 1.6B that is not predicted by the smoothly decreasing loss curve.

The same frozen representations transfer to standard predictive benchmarks: \textsc{Bonbon} achieves state-of-the-art zero-shot binding prediction (MF-PCBA AUPR 0.081, LIT-PCBA-AF3 EF1\% 16.6) and affinity prediction, outperforming both physics-based free energy perturbation and structure-based machine learning models (CASP16 $r = 0.78$, FEP+4 $r = 0.88$, OpenFE $r = 0.76$), at approximately five orders of magnitude greater speed than co-folding methods. In prospective wet-lab validation across nine disease targets spanning seven protein families, \textsc{Bonbon} achieves a 61.5\% dose-response hit rate on de novo compounds, yielding structurally novel molecules not previously reported to inhibit these proteins, including for out-of-distribution targets.

\paragraph{Contributions.}
\begin{itemize}
    \item A foundation model for molecular interactions, extending the foundation-model paradigm to multimodal biology.
    \item Three emergent capabilities arising at hierarchical levels of resolution within a single representation space, without task-specific supervision.
    \item A controlled scaling experiment showing that capabilities remain flat or near baseline through approximately three orders of magnitude of compute and emerge sharply at the largest scale on continuous metrics, while pretraining loss decreases smoothly.
    \item State-of-the-art binding and affinity prediction at five orders of magnitude greater speed than co-folding methods, and prospective wet-lab validation across nine disease targets spanning seven protein families, with a 61.5\% dose-response hit rate on de novo compounds, including for out-of-distribution targets, yielding molecules not previously reported to inhibit these proteins.
    \item A data-heavy scaling regime (1:2{,}000) adopted to accommodate the information sparsity of paired biological data.
\end{itemize}

\subsection{Related Work}
\label{sec:related}

\paragraph{Structure-based interaction modeling.} Co-folding methods ~\citep{abramson_2024_alphafold3, passaro_2025_boltz, chai_2024_chai} predict 3D coordinates of protein-ligand complexes from structure or sequence inputs and have transformed binding pose prediction. However, capabilities beyond pose prediction require supervised heads: across recent structure-prediction models, affinity prediction is a fine-tuned regressor rather than a property recovered from the pretrained representation alone. Generalization is also limited: FoldBench reports substantial memorization on benchmarks with weak performance on unseen ligands~\citep{xu_2025_benchmarking}. In addition, these models systematically misplace allosteric ligands into orthosteric pockets~\citep{nittinger_2025_folding}.

\paragraph{Frozen-embedding interaction models.} Paired protein-ligand encoders trained with contrastive objectives~\citep{singh_2022_learning, gao_2023_drugclip} support binding-prediction retrieval at scale, but their representations have not been shown to learn capabilities beyond the binding signal they were trained on.

\paragraph{Task-specific interaction prediction.} Pocket detection~\citep{leguilloux_2009_fpocket, krivak_2018_p2rank} and allosteric site prediction~\citep{riedlova_2026_protein} have historically been addressed by separately supervised models, each requiring residue-level annotations and 3D protein structures. Structure-based covalent docking tools~\citep{zhu_2014_docking} require the user to specify a priori that a covalent bond forms and predict only the resulting geometry, rather than identifying covalent engagement or the reactive residue.

\paragraph{Emergent capabilities in foundation models.} \citet{wei_2022b_emergent} characterize emergent abilities as tasks not in the pretraining objective that appear with sufficient scale. \citet{schaeffer_2023_emergent} argue that apparent emergence is often a metric artifact: capabilities scale smoothly, and discontinuities reflect threshold-imposing evaluation metrics.

\begin{figure}[!t]
    \centering
    \includegraphics[width=0.90\linewidth]{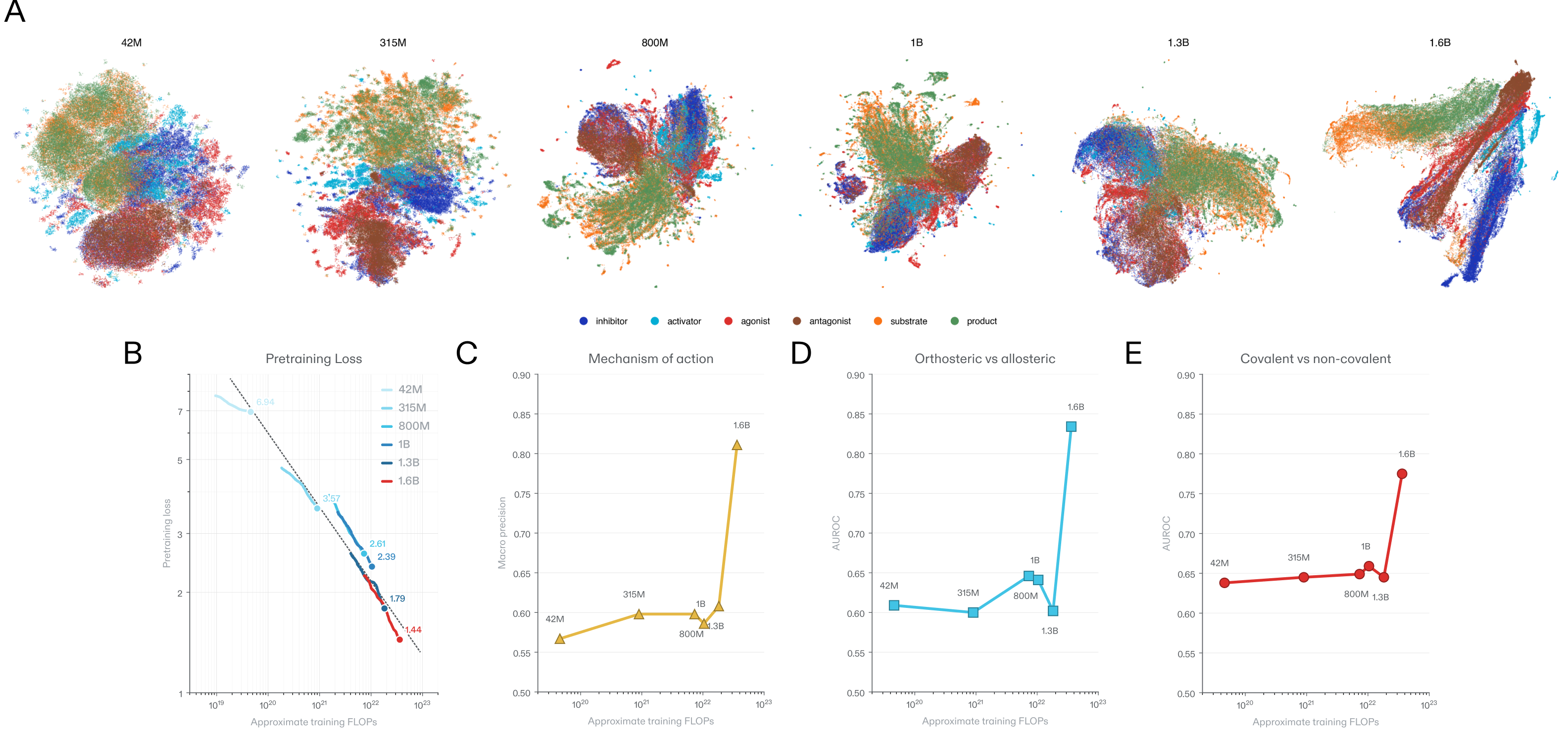}
    \caption{Capability emergence at scale. Controlled scaling experiment across six model scales (42M, 315M, 800M, 1B, 1.3B, 1.6B parameters) under identical conditions except parameter count. Panels (b) to (e) use log FLOPs as the x-axis. \textbf{(a)} UMAP projections (visualization only) of frozen interaction representations at each scale, colored by mechanism-of-action class. \textbf{(b)} Pretraining loss vs.\ training FLOPs. \textbf{(c)} Six-class mechanism-of-action macro precision from a parameter-free signature, mean over ten evaluation splits. \textbf{(d)} Zero-shot orthosteric vs.\ allosteric AUROC on 6{,}032 OOD complexes. Morgan fingerprint baseline shown dashed. \textbf{(e)} Zero-shot covalent vs.\ non-covalent AUROC on the OOD complexes in which neither class carries a detectable electrophilic warhead. All capability evaluations use parameter-free signatures derived from frozen representations and applied unchanged across scales.}
    \label{fig:scaling}
\end{figure}

\section{Model}
\label{sec:model}

\subsection{Architecture}
\textsc{Bonbon} is a transformer-based model trained by self-supervised learning on paired protein-ligand sequences. The architecture is designed for the many-to-many relationships inherent in molecular interaction data, with pre-layer normalization, rotary position embeddings, flash attention, and a 65{,}536-token context window sufficient to encode the longest known protein sequences and accommodate multi-component inputs (Appendix Figure~\ref{fig:A1}).

\subsection{Training Data}
Training data comprised of paired protein-ligand sequences from proprietary screening datasets, enriched with synthetic interaction data generated to match the distributional properties of experimentally measured binding data from that corpus. The generation methodology is outside the scope of this paper and is the subject of a separate technical report. Every evaluation in this paper uses real experimental data from public databases and the published literature; no synthetic data appears in any evaluation set. The corpus spans soluble enzymes, membrane receptors, ion channels, and transcription factors, with ligand chemistries ranging from small molecules to peptides and protein-protein interactions.

\textsc{Bonbon} was trained on approximately 3.2 trillion tokens drawn from this corpus. The sampling distribution was designed to ensure diverse representation across both protein and ligand space, attenuating bias toward well-studied therapeutic targets.

Multiple quality filters were applied. PAINS-pattern compounds~\citep{baell_2010_new} and other reactive or aggregator-prone substructures are excluded to avoid pan-assay artifacts. Proteins with disproportionately broad ligand sets were removed to prevent their patterns from dominating representations. Sequence-level redundancy filters were applied to both the protein and ligand sides of the corpus. All proteins sharing greater than 90\% sequence identity with any evaluation target were excluded prior to training as the basis for all OOD evaluation reported in subsequent Sections~\ref{sec:emergent} and~\ref{sec:translate}.

\subsection{Model Training and Scaling Experiment}
\label{sec:scaling}

The 1.6B production model uses 3.2T training tokens, a 1:2{,}000 parameter-to-token ratio. Hardware, optimizer, and training schedule are described in Appendix~\ref{app:A1-1}.

We chose this data-heavy ratio to accommodate the information sparsity of paired interaction data. To characterize how capabilities relate to scale, we trained six model scales spanning approximately three orders of magnitude in compute (42M, 315M, 800M, 1B, 1.3B, 1.6B parameters). We then evaluated each scale on (i)~pretraining loss, (ii)~zero-shot mechanism-of-action macro precision, (iii)~zero-shot orthosteric/allosteric AUROC on OOD complexes, and (iv)~zero-shot covalent vs.\ non-covalent AUROC on OOD complexes, with and without a detectable electrophilic warhead. Both AUROC evaluations use the parameter-free signatures of Sections~\ref{sec:site} and~\ref{sec:covalent}, with signature features and signs derived once and applied unchanged across all six scales, isolating capability emergence from any signature-tuning effect. All models were trained on the same corpus, sub-sampled to roughly preserve the data-heavy regime, with the same sampling procedure drawing from the full corpus at every scale, so the data mixture, including the proportion of real and synthetic data, is preserved. The architecture family, optimizer configuration and warmup schedule match the 1.6B model, and only the parameter count and its corresponding token budget differ. Each run trained on a predetermined sample from the corpus, with no early stopping or convergence-based termination, and every capability was evaluated with the same protocol on the same held-out sets at every scale. Each scale is a single training run. Training-seed variability was not measured.

Pretraining loss decreases monotonically and continuously with compute, falling from 6.94 at 42M to 1.44 at 1.6B (Figure~\ref{fig:scaling}b). Each of the zero-shot capabilities shows a sharp jump at 1.6B, with the largest absolute increase concentrated at the final scale across all three. All capabilities are absent at the five sub-1.6B scales despite identical training conditions, confirming they are a property of the largest scale rather than of the corpus or training procedure alone. The capabilities show qualitatively different behavior: MoA macro precision stays between 0.567 and 0.608 across the five sub-1.6B scales (0.567, 0.598, 0.598, 0.586, 0.608), then jumps to 0.811 at 1.6B (Figure~\ref{fig:scaling}c). Orthosteric/allosteric AUROC stays at or near the Morgan fingerprint baseline (0.589) across the five smaller scales (0.609, 0.600, 0.646, 0.641, 0.602), then jumps to 0.834 (Figure~\ref{fig:scaling}d). Covalent AUROC measured on all molecules moves within a narrow band across the smaller scales (0.645, 0.678, 0.691, 0.675, 0.695) before reaching 0.838 at 1.6B (Appendix~\ref{app:A5-6}). Ablating warhead presence, we evaluated only complexes in which neither class carries a detectable electrophile. AUROC on this subset is flat at 0.638, 0.645, 0.649, 0.659 and 0.645, then jumps to 0.775 at 1.6B (Figure~\ref{fig:scaling}e). All three sub-threshold series are non-monotone. Against log FLOPs their slopes are 0.012 (mechanism of action), 0.008 (orthosteric/allosteric) and 0.005 (covalent) per tenfold increase in compute, against 0.682, 0.780 and 0.437 respectively for the 1.3B to 1.6B step. Each capability is measured on a non-overlapping protein set, ensuring independence across evaluations, and on a continuous metric, addressing the criticism that apparent emergence can be an artifact of discontinuous evaluation metrics~\citep{schaeffer_2023_emergent}.

Loss-based scaling, the standard quantity used to predict compute-optimal training~\citep{hoffmann_2022_training}, is monotone across the six scales and gives no signal that capabilities are about to appear. Selecting a model size on the basis of loss alone would not reveal whether the next compute step crosses a capability threshold; the threshold is visible only when capabilities are measured directly. The 1.6B scale is the smallest configuration in this experiment at which the threshold is crossed. We report where the transition occurs, not why it occurs. A plausible hypothesis consistent with these observations is that recognizing an interaction property requires composing several pieces of information, none of which is a local property of either sequence, following the intuition of \citet{wei_2022b_emergent} about depth sufficient for sequential composition (Appendix~\ref{app:A1-4}).

\begin{figure}[!t]
    \centering
    \includegraphics[width=0.90\linewidth]{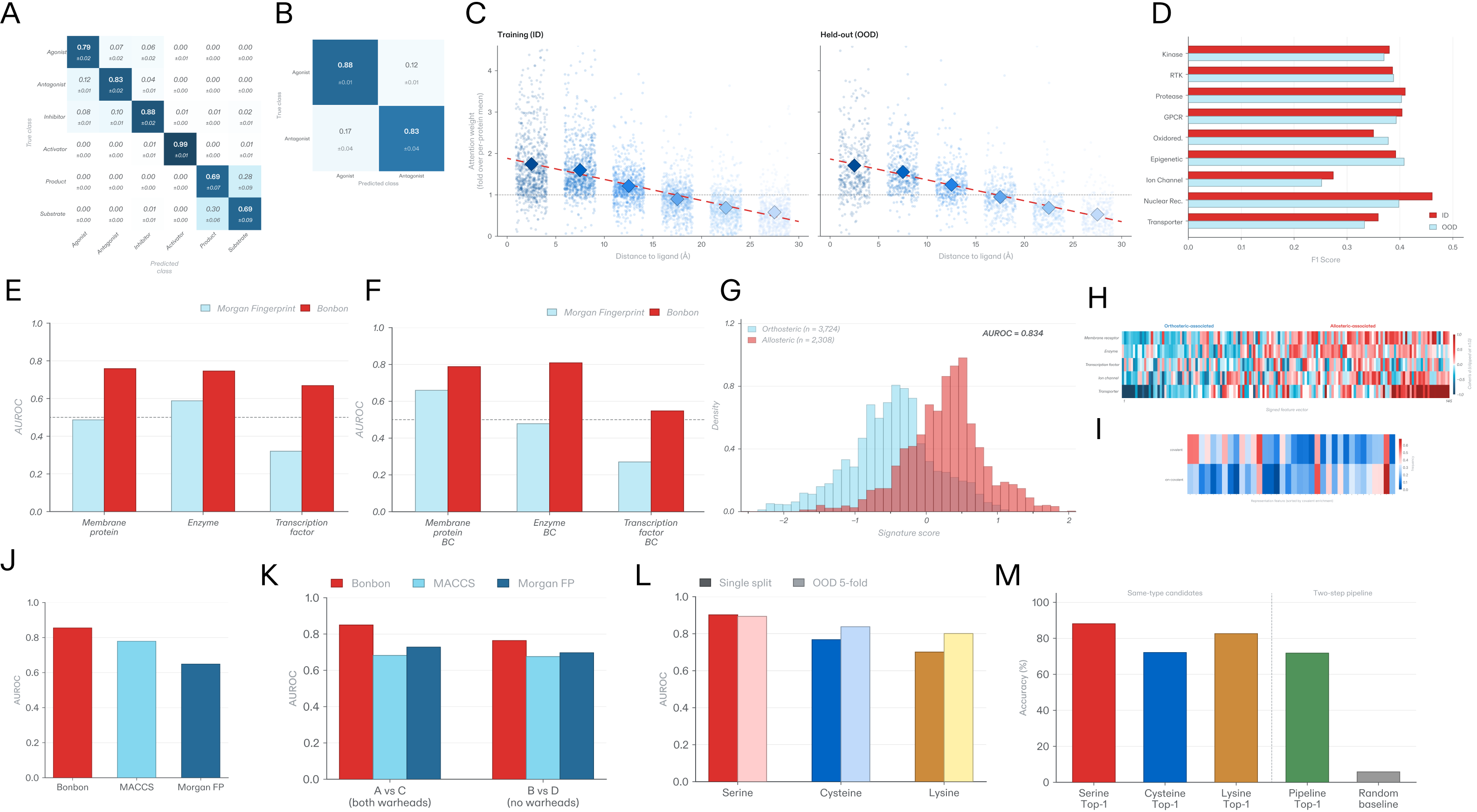}
    \caption{Emergent capabilities across three hierarchical levels, all from frozen representations. \textbf{Mechanism of action:} (a) Six-class signature confusion matrix, column-normalized, mean over ten evaluation splits of a 107{,}027-pair OOD interaction set. (b) Within-receptor agonist/antagonist discrimination across 222 GPCRs. \textbf{Binding site localization and orthosteric/allosteric discrimination:} (c) Per-residue attention vs.\ ligand distance, ID and OOD. (d) Per-family binding site F1, ID and OOD. (e) Orthosteric/allosteric AUROC on 6{,}032 OOD complexes, per family. (f) Within-protein classification on 1{,}860 OOD complexes. (g) Signature score distributions by binding-mode class. (h) Signed feature vector: orthosteric-associated (blue) and allosteric-associated (red) features, by protein family. \textbf{Covalent binding mode:} (i) Activation frequency of the 36 signature features in covalent versus non-covalent complexes, sorted by covalent enrichment. (j) AUROC under full, warhead-matched, and warhead-free conditions. (k) Warhead-controlled comparison. (l) Same-residue-type classification for serine, cysteine, and lysine. (m) Top-1 reactive residue localization accuracy.}
    \label{fig:emergent}
\end{figure}

\section{Emergent Capabilities}
\label{sec:emergent}

\textsc{Bonbon}'s frozen representations encode biological information about molecular interactions at three hierarchical levels of granularity: functional outcome, binding site identity, and bond type. None of these properties is determined by either the protein or the molecule alone, making each a property of the interaction that unimodal models cannot capture. \textsc{Bonbon}'s training data contained no task-specific labels for any of these properties. All three emerge only at the largest model scale (Section~\ref{sec:scaling}).

\subsection{Mechanism of action prediction}
\label{sec:moa}

Mechanism of action describes the functional outcome when a molecule binds a target (e.g., pathway activation vs.\ suppression). UMAP projection of \textsc{Bonbon}'s interaction representations across six mechanism classes (inhibitor, agonist, antagonist, activator, substrate, product) visualizes the hierarchical organization, separating regulatory from metabolic mechanisms (Figure~\ref{fig:scaling}a). A 160-feature parameter-free signature (107{,}027 interactions, molecules below 0.60 Tanimoto similarity to training data; Appendix~\ref{app:A2-1}) yields 81.1\% macro precision.

To eliminate the confound that the model exploits protein-family shortcuts, we evaluated within-receptor agonist/antagonist discrimination on 28{,}126 interactions across 222 GPCRs where every receptor has both classes (Appendix~\ref{app:A2-2}). Holding the receptor constant acts as an ablation for protein identity as a possible shortcut for agonist/antagonist discrimination. \textsc{Bonbon} achieves 85.5\% balanced accuracy (Figure~\ref{fig:emergent}b). No prior computational method has demonstrated zero-shot mechanism-of-action prediction across protein families.

\subsection{Binding site localization and orthosteric/allosteric discrimination}
\label{sec:site}

Having demonstrated functional-level information, we tested whether \textsc{Bonbon} also encodes where on the protein binding occurs, despite no residue-level annotations and no structural information in the training data. \textsc{Bonbon}'s attention weights correlate linearly with ligand proximity (Figure~\ref{fig:emergent}c). Across 1{,}176 co-crystal structures spanning nine protein families (Appendix~\ref{app:A3}), approximately 25\% of attention heads self-organize for spatial localization, achieving F1 = 0.381 $\pm$ 0.142 (in-distribution, ID) and 0.388 $\pm$ 0.124 (OOD), consistent across families including nuclear receptors (F1 = 0.398 OOD) and GPCRs (0.393 OOD), where co-folding methods are known to place ligands unreliably~\citep{zhang_2026_generalization} (Appendix Table~\ref{tab:A5}). Four families perform comparably or better on OOD than on training proteins, ruling out overfitting. On COACH420~\citep{krivak_2018_p2rank}, \textsc{Bonbon} achieves F1 = 0.497, outperforming all three baselines (Appendix Table~\ref{tab:coach420}), while using sequence alone, without the residue-level supervision, 3D coordinates, or pocket annotations the baselines require. The 1{,}176-structure set described above is a harder test than COACH420, spanning targets such as GPCRs and transporters, whose binding sites are less well defined than the enzyme pockets that dominate COACH420.

We next tested whether \textsc{Bonbon} distinguishes orthosteric from allosteric binding. Orthosteric ligands occupy the primary endogenous binding site; allosteric modulators bind secondary pockets that offer higher target specificity. From \textsc{Bonbon}'s frozen interaction representations, we identified a 145-feature parameter-free signature whose discriminative direction between orthosteric and allosteric complexes is consistent across protein families (Appendix~\ref{app:A4-2}).

Across 6{,}032 OOD complexes (141 OOD proteins), the signature achieves AUROC 0.834 versus 0.589 for Morgan fingerprints (Figure~\ref{fig:emergent}e). On the harder task of classifying binding mode among ligands of the same target (where proteins with both orthosteric and allosteric ligands act as an ablation of using protein identity as a shortcut), \textsc{Bonbon} achieves AUROC 0.821 across 1{,}860 OOD complexes (Figure~\ref{fig:emergent}f); on OOD membrane receptors, 0.788. Transcription factors are near chance (0.548), consistent with their smaller training representation. To test whether the site-detection heads causally support this discrimination, we ablated them at inference time and re-ran the same OOD evaluation. Ablating the 25\% site-detection heads drops orthosteric/allosteric AUROC from 0.834 to 0.751; ablating 25\% of heads at random gives 0.816 $\pm$ 0.002 across 10 seeds, so removing the site-detection heads costs roughly 4.5 times what removing an arbitrary subset of the same size costs. The capability degrades but does not disappear: at 0.751 the ablated model remains well above the Morgan fingerprint baseline of 0.589 (Appendix~\ref{app:A4-5}). No prior computational method has demonstrated orthosteric/allosteric discrimination per ligand.

\subsection{Covalent binding mode recognition}
\label{sec:covalent}

Binding site localization identifies where a molecule binds; a finer question is \emph{how}: whether engagement involves covalent bond formation. Covalent drugs can reach targets considered undruggable. Using a 36-feature parameter-free signature from frozen representations, \textsc{Bonbon} achieves AUROC 0.838 on 1{,}212 OOD pairs spanning 237 proteins (Figure~\ref{fig:emergent}j). Two ablations remove warhead presence as an explanation. Among molecules that all carry electrophilic warheads, \textsc{Bonbon} separates those that form covalent bonds from those that do not (AUROC 0.834); among molecules that carry none, it still identifies the covalent binders, whose bonds form through the protein's catalytic machinery rather than ligand reactivity (AUROC 0.775; Appendix~\ref{app:A5-2}). Per-residue representations distinguish the reactive residue from non-reactive same-type residues (serine AUROC 0.902, cysteine 0.817, lysine 0.701), persisting after distance matching (Figure~\ref{fig:emergent}l; Appendix~\ref{app:A5-5}). A two-step pipeline that predicts amino-acid type from pair-level features then ranks pocket residues of the predicted type achieves 71.8\% Top-1 localization accuracy without prior knowledge of the reactive residue (Figure~\ref{fig:emergent}m), a 12-fold improvement over the 5.8\% chance baseline. No prior computational method has demonstrated covalent engagement prediction from sequence alone.

\section{Capabilities Translate to Predictive Accuracy and Novel Inhibitors}
\label{sec:translate}

The three emergent capabilities (Section~\ref{sec:emergent}) demonstrate that \textsc{Bonbon}'s representations encode interaction-level biological information without task-specific training. Two further evaluations test whether this information translates to quantitative predictive accuracy on standard benchmarks (Section~\ref{sec:bench}) and to biochemical activity validated in prospective wet-lab experiments (Section~\ref{sec:lab}).

\subsection{Binding and affinity prediction at five orders of magnitude greater speed}
\label{sec:bench}

For binary binding prediction we use frozen embeddings directly. For binding affinity prediction we train a two-layer MLP probe~\citep{alain_2016_understanding} for ${\sim}15$ minutes on frozen embeddings using affinity measurements from PubChem~\citep{kim_2023_pubchem}, ChEMBL~\citep{zdrazil_2024_chembl}, and BindingDB~\citep{gilson_2015_bindingdb}. Decontamination protocol and dataset preparation are described in Appendix~\ref{app:A6}.

On MF-PCBA, \textsc{Bonbon} achieves state-of-the-art AUPR 0.081 and EF 0.5\% of 24.8 (Figure~\ref{fig:bench}b,c; full per-method comparisons in Appendix Tables~\ref{tab:A12}--\ref{tab:A14}). On LIT-PCBA, \textsc{Bonbon} achieves state-of-the-art EF1\% 11.4, more than double the previous best (Appendix Table~\ref{tab:A12}). On LIT-PCBA-AF3~\citep{shen_2025_unlocking}, a five-target subset of LIT-PCBA restricted to libraries of fewer than 100{,}000 compounds and therefore tractable for co-folding methods, \textsc{Bonbon} achieves state-of-the-art EF1\% 16.6, against 9.3 for the strongest co-folding baseline (Figure~\ref{fig:bench}d; Appendix Table~\ref{tab:litpcba-af3}). On affinity prediction, \textsc{Bonbon}'s MLP probe achieves Pearson $r = 0.78$ on CASP16, $r = 0.88$ on FEP+4, and $r = 0.76$ on OpenFE, the highest reported value on each of the three benchmarks, outperforming both physics-based free energy perturbation and the strongest structure-based models (Figure~\ref{fig:bench}a).

Accuracy is accompanied by orders-of-magnitude reduction in compute. \textsc{Bonbon} processes ${\sim}2{,}500$ pairs/s on-the-fly or ${\sim}42{,}500$ pairs/s with cached representations on a single H100, approximately five orders of magnitude faster than Boltz-2's 0.05 pairs/s (Figure~\ref{fig:bench}e). At proteome scale, screening DrugBank against the human proteome (1.66B pairs) completes in 11 hours versus an estimated ${\sim}1{,}100$ GPU-years for Boltz-2 (Appendix Table~\ref{tab:A15}).

\begin{figure}[!t]
    \centering
    \includegraphics[width=0.90\linewidth]{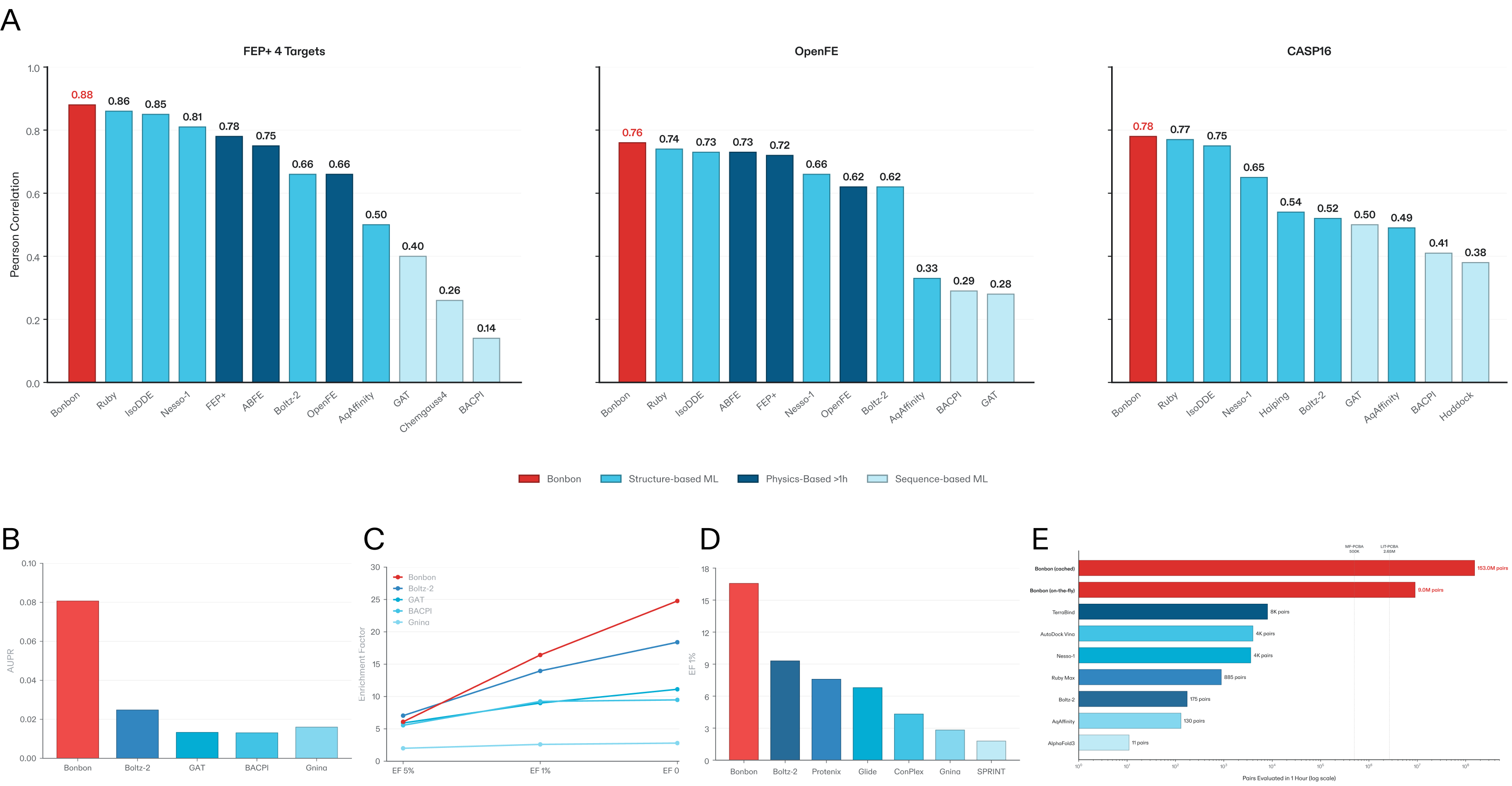}
    \caption{Predictive accuracy and throughput. \textbf{(a)} Binding affinity prediction (Pearson $r$) on CASP16, FEP+4, and OpenFE. \textbf{(b)} MF-PCBA AUPR. \textbf{(c)} MF-PCBA enrichment factor across thresholds. \textbf{(d)} LIT-PCBA-AF3 enrichment factor at 1\%. \textbf{(e)} Inference throughput: pairs evaluated per hour on a single H100, compared to co-folding and docking methods.}
    \label{fig:bench}
\end{figure}

\subsection{Prospective wet-lab experimental validation across nine disease targets}
\label{sec:lab}

\begin{figure}[!t]
    \centering
    \includegraphics[width=0.90\linewidth]{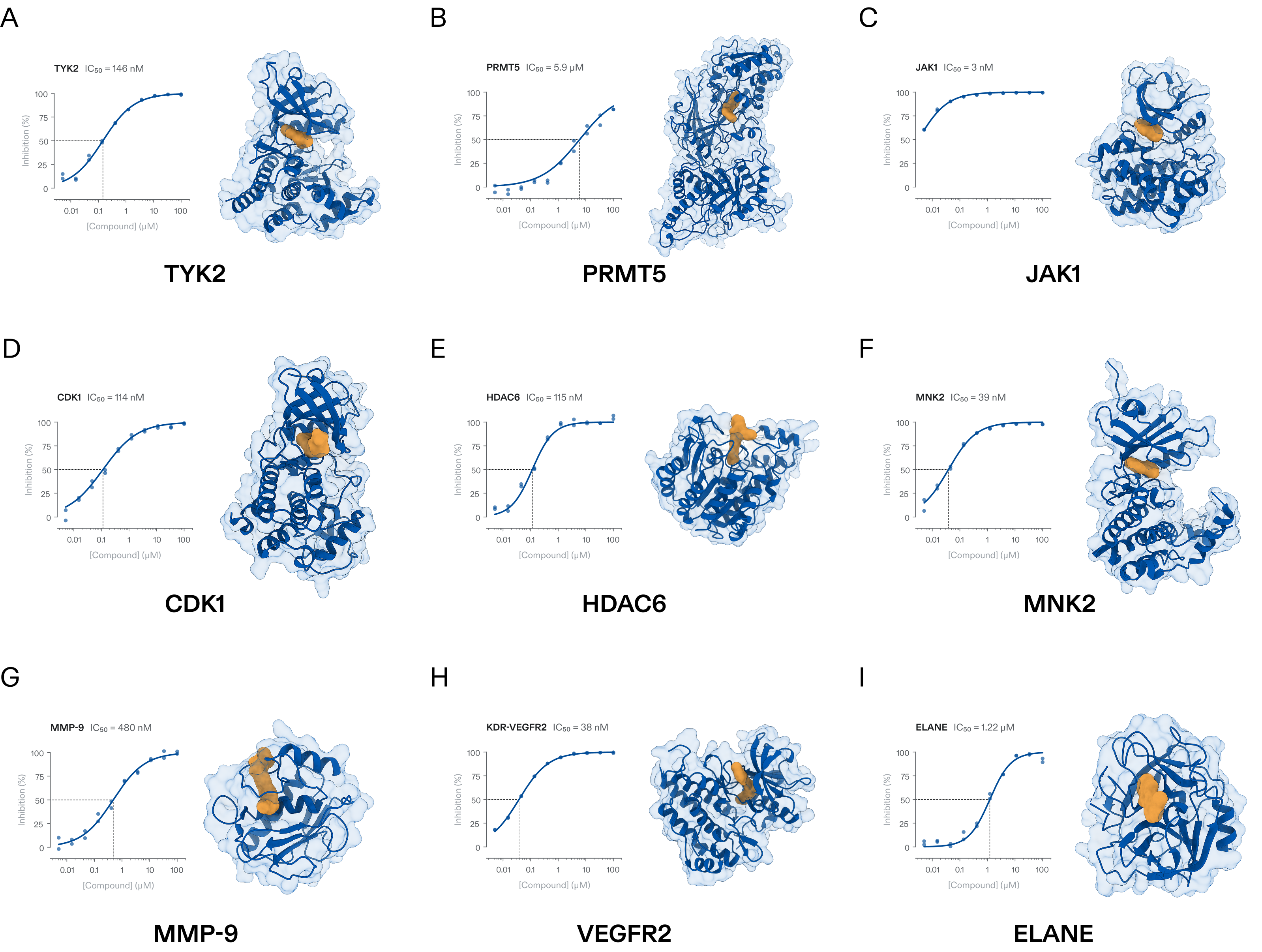}
    \caption{Prospective wet-lab validation across nine disease targets spanning seven protein families. For each target, a ten-point dose-response curve with fitted IC$_{50}$ for one confirmed inhibitor (left) and that compound bound in the target (right; protein ribbon within a transparent surface, compound in orange; Appendix Table~\ref{tab:A7-targets}).}
    \label{fig:lab}
\end{figure}

We used \textsc{Bonbon} to rank molecules generated de novo under synthetic-accessibility constraints against nine disease targets, and up to 24 compounds per target from the top-ranked set were synthesized and assayed, with dose-response IC$_{50}$ determination for actives (Figure~\ref{fig:lab}; Appendix~\ref{app:A7}). All tested compounds have Tanimoto similarity less than 0.60 to any compound in \textsc{Bonbon}'s training data. Structural novelty is assessed as described (Appendix~\ref{app:A7-3}). The nine targets isolate three progressively harder tests.

\textsc{Bonbon} yielded confirmed inhibitors for all nine targets, with IC$_{50}$ values reaching single-digit nanomolar and, on several targets, chemical matter outside the established inhibitor classes.

\paragraph{Novel and efficient inhibitors on well-characterized targets.} TYK2, PRMT5, JAK1 and CDK1 have rich SAR and known pharmacophore classes; they test whether \textsc{Bonbon} identifies efficient and novel inhibitors within established chemical space. On TYK2, 23 compounds were confirmed as inhibitors in dose-response assays, with IC$_{50}$ values from 146\,nM to 40\,$\mu$M and four below 1\,$\mu$M. None is a single-substituent analog of a known inhibitor and 17 are novel molecules. On PRMT5, 16 compounds were confirmed inhibitors, with IC$_{50}$ values from 186\,nM to 143\,$\mu$M. All 16 sit at 0.479--0.597 Tanimoto to the nearest known active and none is a single-substituent analog of any known inhibitor: \textsc{Bonbon} reached the productive region of PRMT5 chemistry without reproducing any individual compound in it, in a single ranked set and without iterative SAR. On JAK1, 21 compounds were confirmed inhibitors, with IC$_{50}$ values from 3\,nM to 28\,$\mu$M, 13 below 1\,$\mu$M and three below 100\,nM. Four are novel molecules, and single-digit nanomolar potency was reached without optimization. On CDK1, all 22 compounds tested were confirmed inhibitors, with IC$_{50}$ values from 114\,nM to 23\,$\mu$M and eight below 1\,$\mu$M; 10 are novel molecules, at 0.279--0.706 Tanimoto to the nearest known active.

\paragraph{New chemistry beyond established pharmacophores.} HDAC6, MNK2 and MMP-9 have inhibitor literatures confined to a narrow region of chemical space; they test whether the model reaches chemical matter that is genuinely new rather than rewarding familiar pharmacophores. The HDAC inhibitor field has been bottlenecked by hydroxamic acid toxicity for two decades: hydroxamic acid appears in every FDA-approved HDAC inhibitor but causes dose-limiting toxicity that has stalled clinical programs. For HDAC6, 15 compounds were confirmed inhibitors in dose-response assays, with IC$_{50}$ values from 115\,nM to 22\,$\mu$M; 12 of the 15 lack hydroxamic acid entirely, and those same 12 are novel molecules. On MNK2, a kinase with no approved inhibitor and only two disclosed clinical programs, 19 compounds were confirmed inhibitors, with IC$_{50}$ values from 39\,nM to 38\,$\mu$M. Eighteen are novel molecules, at 0.269--0.492 Tanimoto to any known MNK2 active with no single-substituent analog among them, the strongest structural novelty in the panel. On MMP-9, where known inhibitors rely on zinc-chelating warheads and broad-spectrum inhibition has produced dose-limiting musculoskeletal toxicity attributed to cross-reactivity within the 23-member MMP family, three compounds were confirmed inhibitors, from 480\,nM to 1.3\,$\mu$M; none carries a strong zinc-chelating warhead (Appendix~\ref{app:A7-4}).

\paragraph{Out-of-distribution targets.} ELANE and VEGFR2 are absent from \textsc{Bonbon}'s training data entirely, with no protein above 90\% sequence identity in training. On ELANE, a serine protease, \textsc{Bonbon} yielded two confirmed inhibitors at 1.22 and 29.4\,$\mu$M, both at or below 0.500 Tanimoto to the nearest known active and neither a single-substituent analog. On VEGFR2, 10 compounds were confirmed inhibitors in dose-response assays, with IC$_{50}$ values from 38\,nM to 82\,$\mu$M; they sit at 0.429--0.617 Tanimoto to the nearest known active, with one novel molecule by every criterion (Appendix~\ref{app:A7-4}).

Across a target panel spanning seven protein families, including out-of-distribution targets, validated hits include molecules novel against the published inhibitor literature, evidence that the representations encode the functional requirements of target engagement rather than memorized pharmacophore patterns.

\section{Discussion}
\label{sec:discussion}

\textsc{Bonbon} learned to predict the functional outcome when a protein and molecule interact, where on the protein, and how they engage, all within a single representation space. All three capabilities arise from frozen representations with no task-specific labels, and none has been demonstrated by any prior computational method. A controlled scaling experiment shows each capability emerges as a sharp jump at 1.6B on continuous metrics, addressing the metric-artifact concern~\citep{schaeffer_2023_emergent}.

\paragraph{Scaling.} Previously reported foundation models exhibiting emergent capabilities converge on parameter-to-token ratios between 1:2 and 1:100. \textsc{Bonbon}'s 1:2{,}000 ratio reflects the sparsity of paired interaction data: the distinction between an agonist and an antagonist, for example, is distributed across millions of examples rather than concentrated in individual sequences, so we allocate far more data per parameter than language modeling conventionally requires. We arrived at this allocation through internal experimentation across token budgets rather than a fixed-compute sweep. A conventional IsoFLOP sweep optimizes pretraining loss~\citep{hoffmann_2022_training}, which is the quantity our scaling results indicate does not predict capability availability (Appendix~\ref{app:A1-4}).

\paragraph{Implications for drug discovery.} Each capability addresses an unmet need. Mechanism-of-action filtering before synthesis replaces costly functional assays. Orthosteric/allosteric discrimination per ligand is a prerequisite for designing allosteric modulators, which offer higher selectivity and access to targets considered undruggable at the orthosteric site. Covalent bond prediction enables both deliberate covalent design and early detection of off-target reactivity currently identifiable only through late-stage experimental assays.

These capabilities and the benchmarks arise from the same frozen representations in a single forward pass. \textsc{Bonbon} provides a computational foundation for drug discovery that replaces sequential, fragmented, structure-dependent tools and laboratory-only measurements with a single representation space, operating at up to 42{,}500 pairs per second. The practical consequence is a shorter design-make-test cycle: interaction properties that today require co-crystal structures, functional assays, or iterative SAR are available computationally before a molecule is synthesized, enabling the evaluation of more hypotheses and more ambitious ones. Appendix~\ref{app:moiety} reports an additional analysis at atomic resolution, identifying which atoms within a ligand drive its activity. The prospective wet-lab validation across nine targets confirms this translates to practice, including validated hits that are novel against the published inhibitor literature, and generalization to out-of-distribution proteins. 

\paragraph{Limitations.} The capabilities reported here resolve coarse distinctions; finer granularity and explicit 3D coordinate extraction remain for future work. All results here are obtained from frozen representations: \textsc{Bonbon} has not been fine-tuned on any downstream task, so what we report is a lower bound on what these representations encode. Additional interaction properties such as residence time, dissociation kinetics, and how conformation and function change after binding are not explored in this work.

\paragraph{Conclusion.} \textsc{Bonbon} demonstrates that the functional grammar of molecular interactions is encoded in the sequences of interacting molecules and recoverable at sufficient scale, without task-specific labels or structural coordinates. The capabilities generalize to out-of-distribution complexes and translate to lab-validated novel inhibitors across nine disease targets. \textsc{Bonbon}'s representation space and the speed at which it operates open directions beyond pairwise prediction, such as generative molecular design conditioned on mechanistic intent, where mechanism, binding site, and binding mode constraints are enforced during generation rather than screened after synthesis; and systems-level interaction profiling, where the computational cost of characterizing every protein-ligand pair in a pathway becomes negligible, enabling the design of molecules with predetermined multi-target profiles and explicit selectivity across protein families. Characterizing not just whether molecules interact but how, at this scale, is a foundation for understanding biological systems.

\paragraph{Safety and ethics.} A model that predicts interaction properties, covalent engagement, and mechanism of action could in principle guide the design of molecules with harmful biological activity. We mitigate this by not releasing model weights or training data publicly; access for academic, non-commercial research is granted case by case. \textsc{Bonbon} predicts interaction properties, not organismal toxicity; translating any prediction to a functional molecule requires synthesis, formulation, and safety testing outside the model's scope.

\begin{ack}
We would like to thank Itsik Pe'er and Pranam Chatterjee for insightful discussions and valuable feedback on the manuscript. We are especially grateful to Mirella Mashiach for her thoughtful guidance and feedback throughout this work.
\end{ack}

\bibliographystyle{unsrtnat}
\bibliography{references}

\newpage
\clearpage
\appendix
\setcounter{figure}{0}
\setcounter{table}{0}
\renewcommand{\thefigure}{A\arabic{figure}}
\renewcommand{\thetable}{A\arabic{table}}
\section*{Appendix}
\setcounter{tocdepth}{2}
\tableofcontents
\clearpage

\noindent This appendix provides supplementary methodological and experimental detail referenced throughout the main text. Sections are lettered A through H in the order they are first referenced from the main text. All figures and tables in this appendix are prefixed ``A'' (e.g., Table~A1, Figure~A1) and are numbered sequentially within the appendix.

\section{Model architecture, training, and scaling experiment}
\label{app:A1}

\subsection{Hardware, optimizer, and training schedule}
\label{app:A1-1}
\textsc{Bonbon} was trained on a cluster of NVIDIA H100 GPUs using PyTorch~2.0 with Fully Sharded Data Parallel (FSDP). Optimization used AdamW with a learning rate of $1{\times}10^{-3}$ and 10{,}000 warmup steps. The model uses pre-layer normalization, rotary position embeddings, flash attention, and a 65{,}536-token context window.

\begin{figure}[H]
    \centering
    \includegraphics[width=\linewidth]{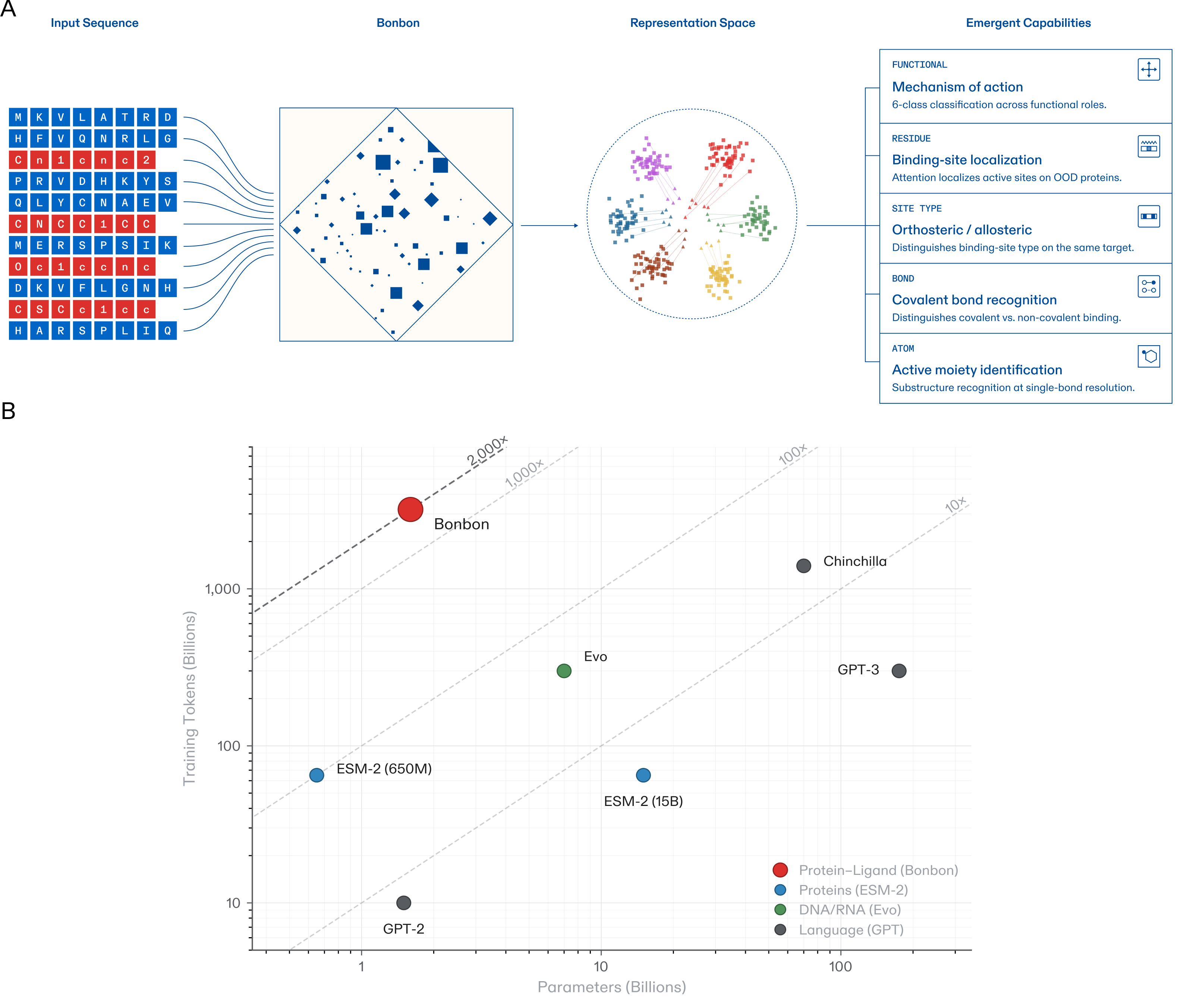}
    \caption{\textsc{Bonbon} design and scale. \textbf{(a)} Schematic of the input-to-prediction pipeline: paired protein and ligand sequences are jointly encoded by a transformer trained with a self-supervised objective, yielding the interaction representations from which the zero-shot capabilities of Section~\ref{sec:emergent} are read out. \textbf{(b)} Parameter-to-token scatter for representative foundation models. \textsc{Bonbon} sits in a data-heavy regime relative to Chinchilla-optimal scaling. Reference table: Table~\ref{tab:A1}.}
    \label{fig:A1}
\end{figure}

\subsection{Training-scale reference table}
\label{app:A1-2}
Table~\ref{tab:A1} lists training scale and compute for representative foundation models in language and biology, together with \textsc{Bonbon}. Figure~\ref{fig:A1}b plots the parameter-to-token ratios from this table.

\begin{table}[H]
\centering
\caption{Training scale of \textsc{Bonbon} compared to representative foundation models. \textsc{Bonbon}'s 1:2{,}000 parameter-to-token ratio places it in a data-heavy regime distinct from language and unimodal biology models.}
\label{tab:A1}
\small
\setlength{\abovecaptionskip}{4pt}
\setlength{\belowcaptionskip}{2pt}
\resizebox{\linewidth}{!}{%
\begin{tabular}{lllllll}
\toprule
Model & Domain & Params & Tokens & Param:Token & FLOPs & Emergent capabilities \\
\midrule
GPT-2 & Language & 1.5B & 10B & 1:7 & $9{\times}10^{19}$ & Coherent generation \\
ESM-2 & Proteins & 650M & 65B & 1:100 & $2.5{\times}10^{20}$ & Weak structure \\
ESM-2 & Proteins & 15B & 65B & 1:4 & $5.8{\times}10^{21}$ & Structure prediction \\
Chinchilla & Language & 70B & 1.4T & 1:20 & $5.9{\times}10^{23}$ & Few-shot reasoning \\
GPT-3 & Language & 175B & 300B & 1:2 & $3.1{\times}10^{23}$ & In-context learning \\
Evo & DNA/RNA & 7B & 300B & 1:43 & $1.3{\times}10^{22}$ & Gene function \\
\textbf{\textsc{Bonbon}} & \textbf{Protein-ligand} & \textbf{1.6B} & \textbf{3.2T} & \textbf{1:2{,}000} & $\mathbf{3.6{\times}10^{22}}$ & \textbf{Interaction biology} \\
\bottomrule
\end{tabular}%
}
\end{table}

\subsection{Scaling experiment configurations}
\label{app:A1-3}
Figure~\ref{fig:scaling} in the main text reports a controlled scaling experiment across six model scales: 42M, 315M, 800M, 1B, 1.3B, and 1.6B parameters. All six models were trained on the same corpus, sub-sampled to roughly preserve the data-heavy regime. Each model uses the same architecture family, optimizer configuration, and warmup schedule as the production 1.6B model; only the parameter count and corresponding token budget differ.

For each scale we evaluated:
\begin{itemize}
    \item \textbf{Pretraining loss vs.\ compute} (Figure~\ref{fig:scaling}b): logged during training and plotted against approximate FLOPs.
    \item \textbf{Six-class mechanism-of-action macro precision} (Figure~\ref{fig:scaling}c): computed on the same evaluation set as Section~\ref{sec:moa} / Appendix~\ref{app:A2-1} (107{,}027 interactions, molecules split at 0.60 Tanimoto, 160-feature parameter-free signature on frozen representations). Every scale is evaluated on the same ten train/test splits of this set and we report the mean across them. These splits resample the evaluation partition only; they do not resample model training.
    \item \textbf{Orthosteric vs.\ allosteric AUROC} (Figure~\ref{fig:scaling}d): computed using the 145-feature parameter-free signature derived from Section~\ref{sec:site} / Appendix~\ref{app:A4}, evaluated on the same 6{,}032 OOD complexes from 141 OOD proteins.
    \item \textbf{Covalent vs.\ non-covalent AUROC} (Figure~\ref{fig:scaling}e): computed using the 36-feature parameter-free signature derived from Appendix~\ref{app:A5}, evaluated on the subset of the 1{,}212 OOD pairs from 237 OOD proteins in which neither class carries a detectable electrophilic warhead (Appendix~\ref{app:A5-2}).
\end{itemize}

The signature features and signs for all three scoring rules, mechanism of action, orthosteric/allosteric and covalent, are derived once from in-distribution data at the 1.6B scale and applied unchanged to all six model scales; this isolates capability emergence from any signature-tuning effect. The orthosteric/allosteric and covalent OOD protein sets share no proteins.

\subsection{On the mechanism of emergence}
\label{app:A1-4}

The mechanism of emergence is an open question in the field. \citet{wei_2022b_emergent} note that few compelling explanations exist for why capabilities emerge in the way they do, and offer two intuitions rather than accounts: depth sufficient for sequential composition, and capacity sufficient for tasks requiring knowledge memorization. Our primary hypothesis is that emergence depends on data and computation.

First, the property has to be representable. Whether a bond forms depends on the reactivity of the ligand's electrophile and the environment of the candidate residue, not on proximity. Two molecules can occupy the same pocket in similar poses while producing opposite functional outcomes. Neither distinction is recoverable from coordinates alone, so a representation organized around geometry cannot hold it. Training on multimodal sequences imposes no such constraint, leaving the model free to learn the interaction pattern directly, putting the capability within reach.

Second, the corpus has to be large enough, also considering patterns are not present in every single pair. For example, whether a given electrophile reacts with a given nucleophile in a given environment is a regularity that exists only across many pairs sharing a protein or a chemotype. This is the reason for training at 1:2{,}000 rather than near the Chinchilla-optimal ratio.

Third, the pattern has to be computable. Recognizing bond formation, for example, requires locating the reactive group on the ligand, identifying candidate nucleophilic residues, assessing the local environment of each, and judging whether that specific combination reacts. These are several composed steps, and none of them is a local property of either sequence. Following Wei's intuition about depth sufficient for sequential composition, capacity below threshold may support the individual steps while the composition remains out of reach.

Our explanation is a hypothesis grounded in the nature of molecular interactions. It is consistent with the observations reported in Section~\ref{sec:scaling} and is not an experimentally established causal mechanism. We view our explanation as another data point in the open debate of explaining the mechanism of emergence.

\section{Mechanism of action: evaluation datasets}
\label{app:A2}

\subsection{Six-class mechanism dataset}
\label{app:A2-1}
We curated 107{,}027 protein-ligand interactions with literature-derived mechanism labels across six classes (inhibitor, agonist, antagonist, activator, substrate, product). \textsc{Bonbon} was trained without any of these labels; the dataset is used exclusively for evaluating whether the embedding space organizes by mechanism. To prevent scaffold memorization in the evaluation, molecules were clustered at 0.60 Tanimoto similarity using Morgan fingerprints (radius 2, 2048 bits), and entire clusters were assigned to either the evaluation-train split or the test split.

\paragraph{Class composition.} The six classes are not represented uniformly across pharmacology databases: regulatory categories (inhibitor, agonist, antagonist, activator) are over-represented relative to metabolic categories (substrate, product), reflecting the focus of historical pharmaceutical research on small-molecule modulators of regulatory targets. Aggregate metrics are reported as macro-averaged precision (unweighted mean across the six classes) to prevent the larger inhibitor class from dominating the score. Because entire Tanimoto clusters are assigned to one side of the split, the held-out fraction varies by class; Table~\ref{tab:A2} reports train and test sizes as means over the ten splits.

\begin{table}[H]
\centering
\caption{Six-class mechanism dataset composition. Train and test sizes are means over the ten evaluation splits.}
\label{tab:A2}
\small
\begin{tabular}{lrrr}
\toprule
Class & Train & Test & Total \\
\midrule
Inhibitor & 27{,}050 $\pm$ 99 & 2{,}950 $\pm$ 99 & 30{,}000 \\
Substrate & 22{,}905 $\pm$ 318 & 1{,}144 $\pm$ 318 & 24{,}049 \\
Product & 21{,}244 $\pm$ 265 & 1{,}078 $\pm$ 265 & 22{,}322 \\
Agonist & 12{,}144 $\pm$ 116 & 991 $\pm$ 116 & 13{,}135 \\
Antagonist & 9{,}218 $\pm$ 108 & 1{,}278 $\pm$ 108 & 10{,}495 \\
Activator & 5{,}506 $\pm$ 68 & 1{,}520 $\pm$ 68 & 7{,}026 \\
Total & 98{,}066 $\pm$ 496 & 8{,}961 $\pm$ 496 & 107{,}027 \\
\bottomrule
\end{tabular}
\end{table}

\paragraph{Signature construction.} The 160-feature signature is a parameter-free scoring rule derived entirely from in-distribution interactions. For each feature in \textsc{Bonbon}'s frozen representations we computed a one-versus-rest effect size for each of the six mechanism classes, and retained the 160 features whose discriminative direction is consistent across protein families rather than driven by any single family. Each retained feature carries a sign per class, taken from the direction of its effect size for that class. Scoring an interaction reduces to one dot product per class between the interaction's representation and that class's signed feature vector, with the predicted class the argmax over the six scores. There are no trainable parameters: feature selection and signs are fixed from in-distribution statistics and applied unchanged to held-out interactions.

\paragraph{Per-class precision.} Table~\ref{tab:A2b} reports signature precision per class as the mean over the ten splits with a 95\% confidence interval on the mean.

\begin{table}[H]
\centering
\caption{Six-class signature precision on held-out interactions, mean over ten evaluation splits.}
\label{tab:A2b}
\small
\begin{tabular}{lccc}
\toprule
Class & Precision & SD & 95\% CI \\
\midrule
Activator & 0.987 & 0.010 & [0.980, 0.994] \\
Inhibitor & 0.879 & 0.020 & [0.865, 0.893] \\
Antagonist & 0.829 & 0.020 & [0.815, 0.843] \\
Agonist & 0.792 & 0.023 & [0.776, 0.808] \\
Substrate & 0.691 & 0.093 & [0.625, 0.758] \\
Product & 0.686 & 0.069 & [0.637, 0.735] \\
\midrule
Macro & 0.811 & 0.010 & [0.803, 0.818] \\
\bottomrule
\end{tabular}
\end{table}

\paragraph{Substrate/product directionality.} Confusion between substrate and product accounts for 97\% and 90\% of each class's total error (Figure~\ref{fig:emergent}a). These two categories differ by reaction direction rather than by chemistry, and they carry the widest intervals in Table~\ref{tab:A2b}.

\subsection{Mechanism overlap at the molecule level}
\label{app:A2-overlap}
Some molecules appear under more than one mechanism, always against different proteins, since each interaction pair is unique. At the molecule level, 1{,}925 of the 58{,}886 molecules in the six-class evaluation set (3.27\%) carry more than one mechanism label. Removing them and repeating the signature classification on the same ten splits gives macro precision of $0.817 \pm 0.017$ against $0.811 \pm 0.010$ on the full set. The paired difference across splits is $+0.007$ with a 95\% confidence interval of $[-0.003, +0.016]$, which contains zero. The six-class result is therefore not an artifact of molecules carrying several labels.

\subsection{GPCR agonist/antagonist evaluation dataset}
\label{app:A2-2}
Table~\ref{tab:A3} gives the composition of the within-receptor evaluation set, and Table~\ref{tab:A4} the resulting agonist versus antagonist discrimination.

\begin{table}[H]
\centering
\caption{GPCR agonist/antagonist evaluation dataset. Every GPCR has both agonist and antagonist ligands; molecule splits ensure structural novelty (Tanimoto $<$ 0.60 to all training molecules).}
\label{tab:A3}
\small
\begin{tabular}{lrrrr}
\toprule
& Train & Test & Val & Total \\
\midrule
Agonist & 12{,}352 & 1{,}543 & 1{,}543 & 15{,}438 \\
Antagonist & 10{,}152 & 1{,}268 & 1{,}268 & 12{,}688 \\
Total & 22{,}504 & 2{,}811 & 2{,}811 & 28{,}126 \\
\bottomrule
\end{tabular}
\end{table}

\begin{table}[H]
\centering
\caption{Signature-based agonist vs.\ antagonist discrimination on GPCRs. Frozen representations; molecule split ensures structural novelty.}
\label{tab:A4}
\small
\begin{tabular}{lc}
\toprule
& \textsc{Bonbon} \\
\midrule
Balanced accuracy & $85.5 \pm 2.1\%$ \\
Agonist precision & $88 \pm 1\%$ \\
Antagonist precision & $83 \pm 4\%$ \\
\bottomrule
\end{tabular}
\end{table}

\section{Binding site localization: dataset, head selection, and per-family results}
\label{app:A3}

\subsection{Dataset construction and methodology}
\label{app:A3-1}
We collected 1{,}176 protein-ligand complexes with experimentally determined co-crystal structures from the PDB, spanning kinases, GPCRs, oxidoreductases, proteases, epigenetic regulators, transporters, ion channels, nuclear receptors, and receptor tyrosine kinases. Targets were selected based on structural availability in the PDB and presence of a co-crystallized drug ligand, without filtering on pocket or cavity characteristics or the existence of prior binding site annotations. Complexes were labeled based on whether the protein appeared in \textsc{Bonbon}'s training data: 782 proteins were found in the training set (in-distribution, ID), while 394 proteins sharing less than 90\% sequence identity with any training protein were designated as the out-of-distribution (OOD) group, representing a stricter test of generalization.

We extracted attention weights from all layers and heads across all 1{,}176 complexes. For each residue, we computed the minimum heavy-atom distance to the ligand and defined binding site residues as those within 8\,\AA{} of any ligand atom. We then measured the correlation between attention weights and inverse ligand distance across all residues.

Attention weights show a monotonic inverse relationship with distance to the nearest ligand atom. Residues within 5\,\AA{} of the ligand receive 1.74$\times$ the per-protein mean attention, while residues beyond 25\,\AA{} receive 0.58$\times$, a 3$\times$ contact-to-distant ratio across the 782 in-distribution proteins. The relationship is strictly monotonic across all six distance bins and consistent across all nine protein families, and it generalizes to the 394 held-out proteins with an equal or larger ratio: contact residues receive 1.71$\times$ the mean and distant residues 0.52$\times$, a 3.3$\times$ ratio, so attention weights reflect learned interaction principles rather than memorization of training examples.

\subsection{Head selection}
\label{app:A3-2}
We identified a subset of attention heads, approximately 25\% of all heads, that show consistently high performance for binding site detection. We refer to these as site-detection heads and use them for all subsequent binding site evaluation. Early-layer heads act as universal binding site detectors and generalist supporting heads, while later-layer heads specialize for particular protein families; for example, certain heads show preferential performance on membrane protein families such as GPCRs and transporters. The emergence of this hierarchical specialization is itself noteworthy: without any residue-level supervision, the model has self-organized a subset of its attention mechanism for spatial localization, with progressive refinement from general to family-specific detection.

\subsection{Per-family binding site prediction}
\label{app:A3-3}
Table~\ref{tab:A5} reports binding site prediction F1 per protein family, in distribution and out of distribution.

\begin{table}[H]
\centering
\caption{Zero-shot binding site prediction across protein families. F1 for residue-level classification using frozen attention weights, mean $\pm$ standard deviation. No residue-level supervision during training. ID = proteins in training set; OOD = proteins held out entirely ($<$ 90\% sequence identity to any training protein). $n$ = number of complexes per family.}
\label{tab:A5}
\small
\begin{tabular}{lccc}
\toprule
Family & $n$ & ID & OOD \\
\midrule
Kinase & 213 & 0.381 $\pm$ 0.090 & 0.370 $\pm$ 0.091 \\
Protease & 180 & 0.410 $\pm$ 0.163 & 0.403 $\pm$ 0.154 \\
GPCR & 156 & 0.404 $\pm$ 0.078 & 0.393 $\pm$ 0.102 \\
Oxidoreductase & 100 & 0.350 $\pm$ 0.155 & 0.378 $\pm$ 0.113 \\
Epigenetic & 61 & 0.392 $\pm$ 0.122 & 0.409 $\pm$ 0.069 \\
Ion channel & 56 & 0.274 $\pm$ 0.131 & 0.252 $\pm$ 0.123 \\
RTK & 38 & 0.386 $\pm$ 0.067 & 0.388 $\pm$ 0.058 \\
Nuclear receptor & 31 & 0.461 $\pm$ 0.128 & 0.398 $\pm$ 0.079 \\
Transporter & 30 & 0.359 $\pm$ 0.176 & 0.333 $\pm$ 0.129 \\
Other & 311 & 0.389 $\pm$ 0.167 & 0.393 $\pm$ 0.142 \\
\midrule
\textbf{Overall} & \textbf{1{,}176} & \textbf{0.381 $\pm$ 0.142} & \textbf{0.388 $\pm$ 0.124} \\
\bottomrule
\end{tabular}
\end{table}

\subsection{COACH420 comparison}
\label{app:A3-4}
Table~\ref{tab:coach420} compares \textsc{Bonbon} against established binding site prediction methods on COACH420.

\begin{table}[H]
\centering
\caption{Binding site prediction on COACH420. F1 for residue-level binding site identification. \textsc{Bonbon} uses sequence alone, with no residue-level supervision, no 3D coordinates and no pocket annotations.}
\label{tab:coach420}
\small
\begin{tabular}{lc}
\toprule
Method & F1 \\
\midrule
\textbf{\textsc{Bonbon}} & \textbf{0.497} \\
Fpocket~\citep{leguilloux_2009_fpocket} & 0.45 \\
P2Rank~\citep{krivak_2018_p2rank} & 0.44 \\
ESMBind~\citep{schreiber_2023_esmbind} & 0.32 \\
\bottomrule
\end{tabular}
\end{table}

\subsection{Structurally challenging targets}
\label{app:A3-5}
\begin{figure}[H]
\centering
\includegraphics[width=\linewidth]{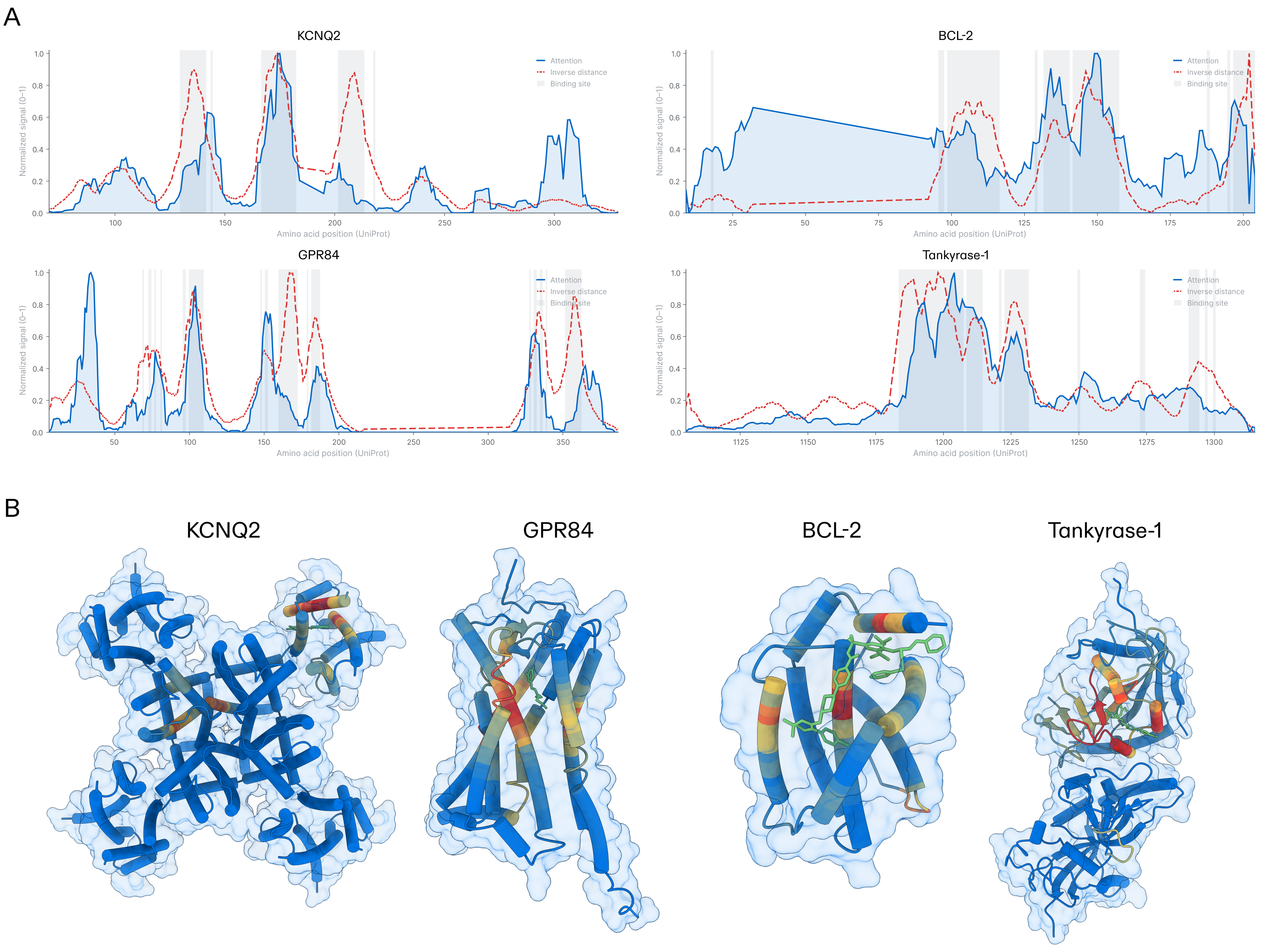}
\caption{\textbf{(a)} Attention overlay traces on four structurally challenging targets. Blue solid: attention signal; red dashed: inverse distance to ligand; grey shading: binding site residues. \textbf{(b)} 3D attention visualization on the same four targets. KCNQ2: attention concentrates on helices surrounding the ligand binding site. BCL-2: attention localizes to the BH3-binding groove on a shallow PPI surface. GPR84: attention maps the orthosteric pocket of an orphan GPCR. Tankyrase-1: attention concentrates on the ligand-binding chain of the homodimer while the non-binding chain remains cold, demonstrating ligand-conditioned localization. \textsc{Bonbon} also distinguishes orthosteric from allosteric binding sites from frozen representations.}
\label{fig:challenging}
\end{figure}

The results above aggregate across many proteins. To examine whether \textsc{Bonbon} localizes binding sites on individual targets where structure-dependent methods face fundamental limitations, we visualized attention patterns on four structurally diverse proteins spanning orphan GPCRs, ion channels, nuclear receptors, and a protein-protein interaction target with a conformationally plastic binding site (Figure~\ref{fig:challenging}). Two representative cases are described below.

\paragraph{BCL-2.} BCL-2 is an anti-apoptotic protein whose drug-binding site is a protein-protein interaction interface, the BH3-binding groove, rather than a conventional pocket. The groove is a shallow, extended hydrophobic surface formed by the BH1, BH2, and BH3 domains~\citep{carogomez_2019_exploring, petros_2001_solution} that lacks the deep pockets typical of small-molecule targets~\citep{verdine_2007_challenge}. The groove exists in the apo state but is conformationally plastic~\citep{rautureau_2010_intrinsically}: an adjacent 58-residue intrinsically disordered flexible loop domain can fold toward the protein core and occlude the BH1--BH3 binding surface~\citep{carogomez_2019_exploring}, and phosphorylation of this loop allosterically modulates groove accessibility from the opposite face of the protein~\citep{lan_2020_anti}. A shallow PPI surface, conformational plasticity, and dynamic occlusion by an adjacent disordered region together make BCL-2 a challenging target for geometry-based methods. \textsc{Bonbon}'s attention concentrates on the BH1 and BH2 domains, which together with BH3 form the hydrophobic groove targeted by BH3-only proteins and small-molecule mimetics such as navitoclax~\citep{carogomez_2019_exploring}. Attention on BH3 is moderate and on BH4 minimal, consistent with its distance from the binding surface. The model identifies the functional binding region from sequence alone, without structural coordinates or pocket geometry.

\paragraph{GPR84.} GPR84 is a GPCR that has been classified as an orphan receptor for over two decades~\citep{luscombe_2020_years, alexander_2025_iuphar}, had no experimentally determined structure until 2023~\citep{liu_2023_structural, zhang_2023_pro}, and features a fully occluded orthosteric pocket that is inaccessible from the extracellular surface~\citep{zhang_2023_pro}. Attention concentrates on residues within 5\,\AA{} of the ligand (Spearman $\rho = +0.558$), correctly localizing the orthosteric binding site across transmembrane helices despite membrane embedding and conformational flexibility.

Two further targets are noted briefly: the potassium voltage-gated channel KCNQ2, where attention localizes the pore-domain binding site ($\rho = +0.526$, OOD), concentrating on residues near the selectivity filter region; and Tankyrase-1, a poly(ADP-ribose) polymerase with 153 intrinsically disordered residues flanking the PARP catalytic domain, where attention concentrates on the catalytic domain binding site ($\rho = +0.570$, ID) while ignoring the extensive disordered regions. Binding site localization therefore extends to protein classes where structural methods face fundamental limitations, including conformationally plastic PPI targets, membrane-embedded receptors, and ion channels, capturing sequence-encoded determinants of ligand engagement rather than geometric pocket features.

\section{Orthosteric/allosteric discrimination: dataset, signature construction, and analysis}
\label{app:A4}

\subsection{Dataset}
\label{app:A4-1}
Allosteric and orthosteric binding annotations were curated from published databases, yielding 19{,}027 protein-ligand complexes across 463 proteins. Complexes were split by protein identity: 322 in-distribution proteins (12{,}995 complexes) for signature derivation, and 141 out-of-distribution proteins (6{,}032 complexes) for evaluation. Both-class proteins (those with at least 2 ligands of each binding type) were used for the within-protein evaluation: 18 OOD proteins (1{,}860 complexes).

\subsection{Signature construction}
\label{app:A4-2}
The 145-feature signature is a parameter-free scoring rule derived entirely from in-distribution complexes. For each feature in \textsc{Bonbon}'s learned representations, we computed the effect size between allosteric and orthosteric complexes and selected 145 features that show consistent discriminative signal across protein families. Each selected feature is assigned a sign from the direction of its effect size. Scoring a new complex reduces to a single dot product between the complex's representation and the signed feature vector. There are no trainable parameters: feature selection and signs are fixed from in-distribution statistics and applied to out-of-distribution complexes without modification.

\subsection{Classification results}
\label{app:A4-3}
Table~\ref{tab:A6} reports zero-shot orthosteric versus allosteric classification against the Morgan fingerprint baseline, overall and per family.

\begin{table}[H]
\centering
\caption{Zero-shot orthosteric vs.\ allosteric classification. 145-feature signature (no trained parameters) on OOD proteins. MFP = Morgan fingerprints. Bold indicates AUROC $>$ 0.6. The TF BC OOD result (0.548) is consistent with the smaller training representation of transcription factors and is reported as a domain coverage limitation.}
\label{tab:A6}
\small
\begin{tabular}{lcccc}
\toprule
Evaluation subset & Proteins & Complexes & Signature & MFP \\
\midrule
All OOD & 141 & 6{,}032 & \textbf{0.834} & 0.589 \\
GPCR OOD & 60 & 3{,}684 & \textbf{0.759} & 0.487 \\
Enzyme OOD & 61 & 1{,}480 & \textbf{0.746} & 0.646 \\
TF OOD & 5 & 445 & \textbf{0.669} & 0.320 \\
Both-class OOD & 18 & 1{,}860 & \textbf{0.821} & 0.609 \\
GPCR BC OOD & 9 & 1{,}229 & \textbf{0.788} & 0.660 \\
Enzyme BC OOD & 5 & 226 & \textbf{0.793} & 0.568 \\
TF BC OOD & 2 & 361 & 0.548 & 0.270 \\
\bottomrule
\end{tabular}
\end{table}

\subsection{Score distribution by family}
\label{app:A4-4}
\begin{figure}[H]
    \centering
    \includegraphics[width=1.0\linewidth]{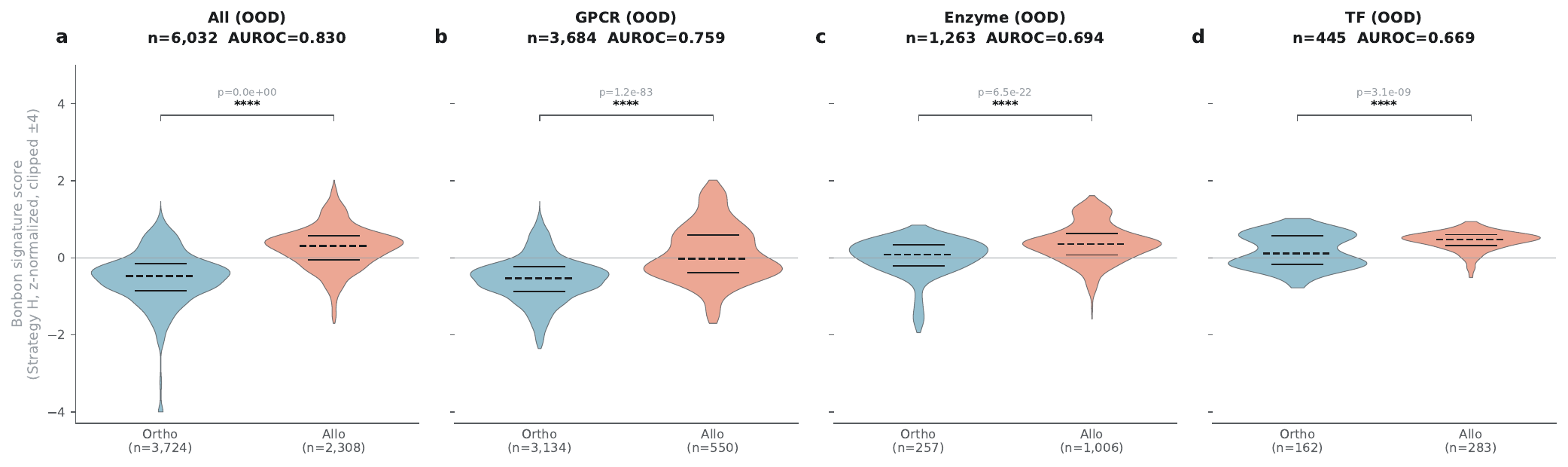}
    \caption{Signature score distributions by binding-mode class (violin plots). Bimodal separation between orthosteric and allosteric classes is visible across membrane receptors and enzymes; on transcription factors, the separation collapses, consistent with the lower AUROC on that family.}
    \label{fig:A2}
\end{figure}

Figure~\ref{fig:A2} shows the distribution of signature scores by protein family underlying the AUROCs in Table~\ref{tab:A6}.

\subsection{Head ablation}
\label{app:A4-5}
\begin{figure}[H]
\centering
\includegraphics[width=0.60\linewidth]{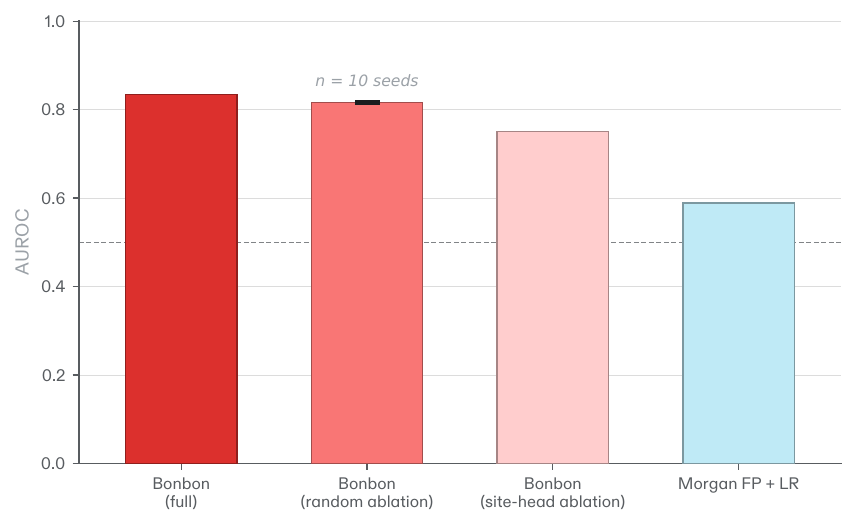}
\caption{Orthosteric/allosteric AUROC after ablating the site-detection heads, compared with random-head controls and the Morgan fingerprint baseline.}
\label{fig:head-ablation}
\end{figure}

Ablation was performed at inference time with no retraining, on the OOD dataset described above. Site-detection heads (25\% of all heads, selected as in Appendix~\ref{app:A3-2}) were zeroed; the control zeroes a random 25\% of heads, repeated over 10 seeds (Figure~\ref{fig:head-ablation}).

The orthosteric/allosteric discrimination capability can be explained as a combination of spatial localization of the binding site from the site-specific heads and binding-mode features related to the interaction representations, with both the heads and the representations contributing to the signature.

\section{Covalent binding mode recognition: dataset, controls, and residue localization}
\label{app:A5}

\subsection{Dataset construction}
\label{app:A5-1}
We curated 1{,}490 protein-ligand pairs (970 covalent, 520 non-covalent) from covalent and non-covalent co-crystal structures in the RCSB PDB, spanning 322 unique proteins. Pocket residues were defined as heavy atoms within 5\,\AA{} of any ligand heavy atom (24{,}811 total). Covalent bond assignments were derived from PDB mmCIF records cross-referenced with the Chemical Component Dictionary. Proteins were split into in-distribution (85 proteins, 278 pairs) and out-of-distribution (237 proteins, 1{,}212 pairs) by sequence identity ($<$ 90\% to training data). All main-text results report OOD performance.

\subsection{Warhead controls}
\label{app:A5-2}
\begin{figure}[H]
\centering
\includegraphics[width=0.62\linewidth]{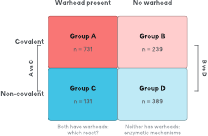}
\caption{The four warhead control groups. Rows give the observed bond type, columns whether the molecule carries a detectable electrophilic warhead. The two controlled comparisons are A vs.\ C, where both groups carry warheads and only one reacts, and B vs.\ D, where neither carries a detectable warhead. Counts match Table~\ref{tab:A7}.}
\label{fig:groups}
\end{figure}

Each molecule was screened for 20 electrophilic substructure patterns (acrylamide, Michael acceptor, nitrile, haloacetamide, epoxide, aldehyde, sulfonyl fluoride, and others) using RDKit SMARTS matching. This yields the four groups in Table~\ref{tab:A7} and Figure~\ref{fig:groups}, which enable two controlled comparisons.

\begin{table}[H]
\centering
\caption{Warhead control groups for covalent binding analysis.}
\label{tab:A7}
\small
\begin{tabular}{llllc}
\toprule
Group & Bond type & Electrophile & $n$ & Description \\
\midrule
A & Covalent & Present & 731 & Warhead engages the target \\
B & Covalent & Absent & 239 & Enzymatic covalent mechanism \\
C & Non-covalent & Present & 131 & Warhead present but does not react \\
D & Non-covalent & Absent & 389 & No electrophilic groups \\
\bottomrule
\end{tabular}
\end{table}

\paragraph{Groups A vs.\ C: both carry reactive groups, only one reacts.}
Both groups carry electrophilic warheads; Group A molecules form covalent
bonds and Group C molecules do not, despite carrying the same types of
reactive substructure. A model that detects warhead presence alone would
score the two groups identically. \textsc{Bonbon} achieves AUROC 0.834
against 0.728 for Morgan fingerprints, and MACCS structural keys, which
reach 0.779 on the full comparison, fall to 0.682 here, the weakest
fingerprint baseline.

\paragraph{Groups B vs.\ D: neither carries a detectable reactive group.}
This comparison removes warhead detection as an explanatory variable
entirely, since covalent binding in Group B proceeds through enzymatic
mechanisms in which the protein's catalytic machinery drives bond formation
rather than the ligand's intrinsic reactivity. \textsc{Bonbon} achieves
AUROC 0.775 against 0.697 for Morgan fingerprints and 0.676 for MACCS, so
discriminative signal persists on molecules carrying no electrophilic
substructure pattern.

\subsection{Classification methodology}
\label{app:A5-3}
\paragraph{Binary covalent vs.\ non-covalent signature (Section~\ref{sec:covalent}).} From frozen interaction representations, we identified 36 features whose discriminative direction between covalent and non-covalent complexes is consistent across protein families on in-distribution data, selected by the same effect-size procedure as the orthosteric/allosteric signature (Appendix~\ref{app:A4-2}), and constructed a parameter-free scoring rule: a single dot product between each complex's representation and the signed feature vector. There are no trained parameters. On the out-of-distribution split the signature reaches F1 0.840 and average precision 0.927, with precision 85.9\% and recall 82.1\%. Per-protein validation was performed on within-protein covalent vs.\ non-covalent pairs from 215 both-class proteins, computing the AUROC per protein using the same signature.

\paragraph{Reactive residue localization.} For per-residue classification of ``is this the covalent residue,'' we trained logistic regression classifiers (L1/L2 regularization, $C = 0.1$, saga solver) on per-residue frozen representations with standardized features.

\subsection{Reactive residue localization}
\label{app:A5-4}
Reactive residue localization was evaluated on 883 OOD covalent complexes.

\paragraph{Two-step pipeline.} Step 1 predicts amino acid type from pair-level features; Step 1 performance is reported in Table~\ref{tab:A9}. Step 2 ranks pocket residues of the predicted type, reaching 71.8\% Top-1 accuracy against a 5.8\% chance baseline, with 84.1\% Top-3 and a mean rank of 2.3.

\begin{table}[H]
\centering
\caption{Step 1 type prediction AUROC (out-of-distribution).}
\label{tab:A9}
\small
\begin{tabular}{lcc}
\toprule
Target residue & AUROC & $n$ positive (OOD) \\
\midrule
Serine & 0.879 & 327 \\
Threonine & 0.861 & 151 \\
Lysine & 0.618 & 60 \\
Cysteine & 0.561 & 149 \\
\bottomrule
\end{tabular}
\end{table}

\subsection{Distance-controlled analysis}
\label{app:A5-5}
To rule out the confound that reactive residues are systematically closer to the ligand, we retained only same-type structural residues within $\pm$0.5\,\AA{} or $\pm$0.25\,\AA{} of any reactive residue's distance. This is an ablation of proximity: distance matching removes distance as an available cue. After distance matching, the signal persists for all three residue types at $\pm$0.5\,\AA{}, and serine and cysteine remain above chance even under strict $\pm$0.25\,\AA{} matching, while lysine's separation is largely proximity-driven. Under five-fold cross-validation on OOD proteins the same discrimination gives 0.894 for serine, 0.838 for cysteine and 0.801 for lysine.

\begin{table}[H]
\centering
\caption{Distance-controlled AUROC for reactive residue identification. Serine and cysteine retain signal well above chance after strict distance matching, confirming the representations encode catalytic-site context beyond proximity. Lysine collapses to chance at the strictest matching, suggesting its signal is primarily distance-driven.}
\label{tab:A11}
\small
\begin{tabular}{lcccc}
\toprule
Residue & Original AUROC & $\pm$0.5\,\AA{} matched & $\pm$0.25\,\AA{} matched & $n$ (original) \\
\midrule
Serine & 0.902 & 0.877 & 0.725 & 412 positive \\
Cysteine & 0.817 & 0.796 & 0.758 & 179 positive \\
Lysine & 0.701 & 0.698 & 0.500 & 72 positive \\
\bottomrule
\end{tabular}
\end{table}

\subsection{Covalent recognition across model scales}
\label{app:A5-6}
\begin{figure}[H]
\centering
\includegraphics[width=0.55\linewidth]{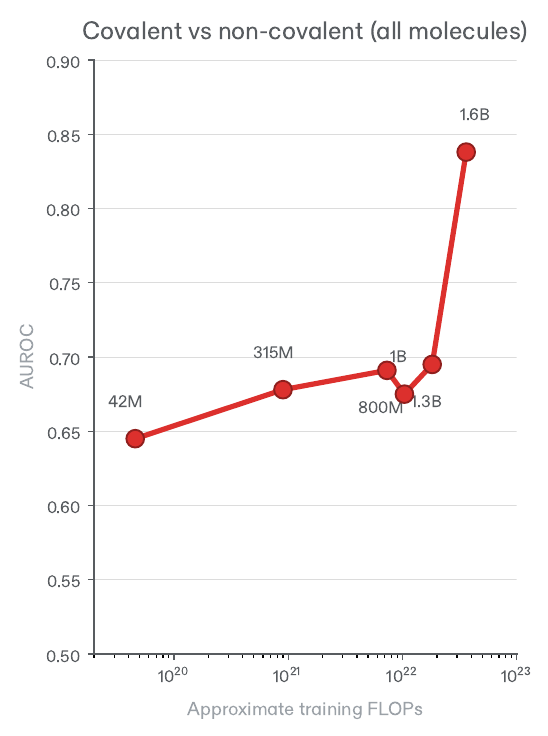}
\caption{Covalent versus non-covalent AUROC across model scales, measured on all molecules.}
\label{fig:covalent-scaling}
\end{figure}

Covalent versus non-covalent AUROC measured on all molecules is 0.645, 0.678, 0.691, 0.675, 0.695, and 0.838 from 42M to 1.6B parameters (Figure~\ref{fig:covalent-scaling}). This contrast includes molecules carrying detectable electrophiles, whose presence is a local property of the ligand and is learnable at smaller scales, which lifts the sub-1.6B values above those of the warhead-free evaluation in Figure~\ref{fig:scaling}e without producing a trend that anticipates the jump at 1.6B.

\section{Active moiety identification: dataset construction, metric, and generalization}
\label{app:moiety}

Active moieties are the chemical substructures responsible for engaging the binding site: a kinase inhibitor's hinge-binding heterocycle is its active moiety, while its solvent-exposed substituents tune selectivity but can be replaced without disrupting binding. Identifying them is a prerequisite for rational optimization, since knowing which atoms to preserve and which to modify is the basis of SAR analysis. Existing approaches require a ligand-bound 3D structure or a docking pose, and docking and co-folding predict contacts rather than which atoms drive activity. No moiety labels appear in \textsc{Bonbon}'s training data.

Ground-truth active moieties were derived by automated matched molecular pair analysis of ChEMBL structure-activity data. For each target with at least 100 active compounds (IC$_{50}$ $<$ 1\,$\mu$M), actives were grouped by Butina clustering at Tanimoto 0.40 on Morgan fingerprints, and matched pairs differing by a single structural transformation were generated within each cluster. Atoms whose single-atom modification shifted activity more than tenfold across at least three pairs were scored as essential and converted to SMARTS patterns; the pipeline was validated against five proteins with expert-curated SMARTS from the medicinal chemistry literature. The resulting dataset spans 145 proteins across five target classes with 241{,}035 protein-ligand pairs.

\begin{figure}[H]
\centering
\includegraphics[width=\linewidth]{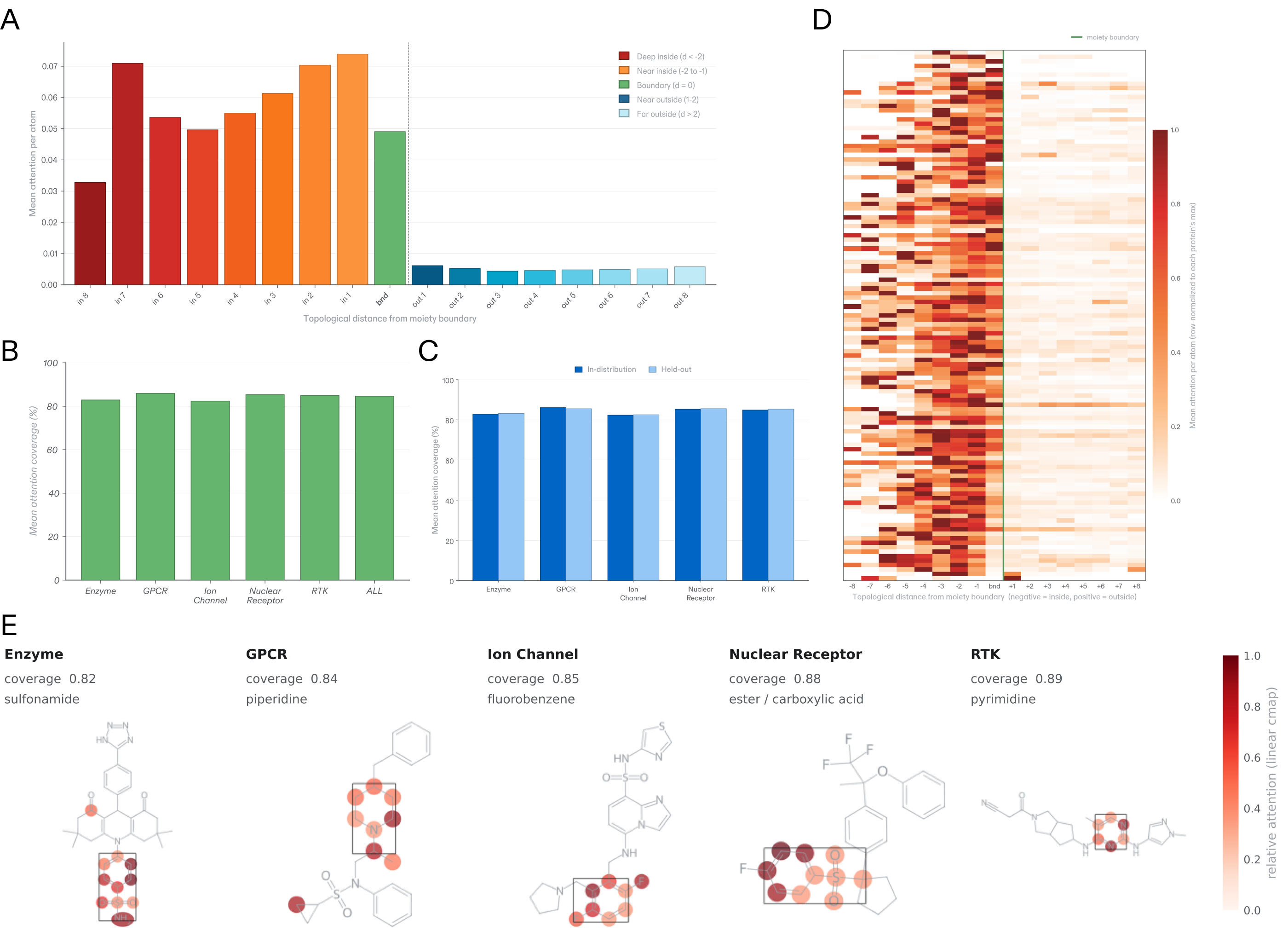}
\caption{\textbf{(a)} Attention boundary profile: mean attention per atom as a function of topological distance from the moiety edge. \textbf{(b)} Mean attention coverage by protein class. \textbf{(c)} Chemotype generalization: mean attention coverage for in-distribution and held-out molecules across protein classes. \textbf{(d)} Per-protein boundary heatmap across 119 proteins sorted by boundary sharpness; the green line marks the moiety edge. \textbf{(e)} Per-atom attention across five protein classes: attention highlights the annotated active moiety (boxed) regardless of protein family or molecular scaffold.}
\label{fig:moiety-panel}
\end{figure}

Attention coverage for a molecule is the fraction of total ligand attention allocated to active moiety atoms, bounded in $[0,1]$. The chance baseline of 0.36 is the average fraction of moiety atoms per molecule, that is, the coverage uniform attention would produce; coverage is bounded and threshold-free, so it imposes no discontinuity on the scaling results~\citep{schaeffer_2023_emergent}. Mean coverage is 0.85 across 241{,}035 molecules and 145 proteins, and is uniform across the five target classes, from 82.4\% for ion channels to 85.9\% for GPCRs, which span different binding site architectures (Table~\ref{tab:moiety}; Figure~\ref{fig:moiety-panel}).

\begin{figure}[H]
\centering
\includegraphics[width=\linewidth]{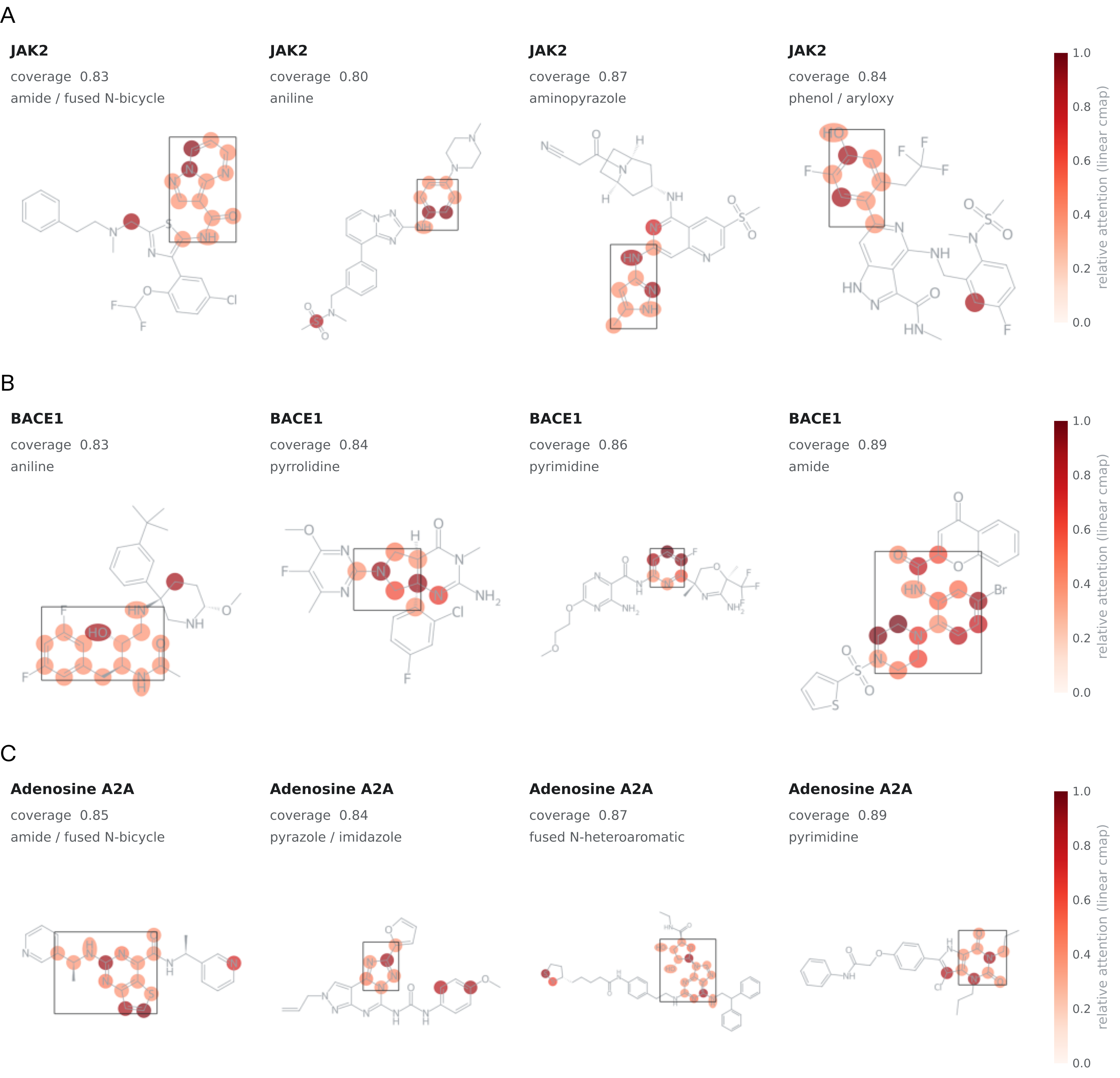}
\caption{\textbf{(a--c)} Interaction-conditioned moiety identification on three out-of-distribution proteins, JAK2, BACE1, and adenosine A2A: four structurally distinct ligands per protein, each with a different active moiety, showing that moiety detection depends on the protein-ligand interaction rather than on the protein alone.}
\label{fig:moiety-per-protein}
\end{figure}

To test generalization to unseen chemical matter, all molecules were clustered into 24{,}525 chemotype groups at Tanimoto 0.60 (Morgan fingerprints, radius 2, 2048 bits) by leader clustering, and 6{,}197 complete clusters were held out, so that no held-out molecule shares greater than 0.60 Tanimoto similarity with any in-distribution molecule. Coverage is 84.6\% in-distribution and 84.7\% across the 60{,}514 held-out molecules, with per-class divergence below 0.7 percentage points. With the protein held constant, attention tracks a different moiety in each of four structurally distinct ligands for JAK2, BACE1 and adenosine A2A, all out of distribution: the pattern depends on the pair, not the protein alone (Figure~\ref{fig:moiety-per-protein}).

Across model scales, mean attention coverage is 0.362 at 42M, 315M, 800M and 1B and 0.365 at 1.3B, against a chance baseline of 0.36, then rises to 0.85 at 1.6B.

\begin{table}[H]
\centering
\caption{Active moiety attention coverage across protein classes. Coverage is the fraction of total ligand attention allocated to active moiety atoms; chance baseline 0.36. No substructure annotations appear in training.}
\label{tab:moiety}
\small
\begin{tabular}{lrrr}
\toprule
Target class & Proteins & Molecules & Mean coverage \\
\midrule
Enzyme & 7 & 31{,}159 & 82.9\% \\
GPCR & 36 & 80{,}270 & 85.9\% \\
Ion channel & 37 & 40{,}584 & 82.4\% \\
Nuclear receptor & 36 & 26{,}312 & 85.3\% \\
RTK & 29 & 62{,}710 & 85.0\% \\
\textbf{All} & \textbf{145} & \textbf{241{,}035} & \textbf{84.6\%} \\
\bottomrule
\end{tabular}
\end{table}

\paragraph{Attention at the moiety boundary.} Coverage measures how much attention falls inside the moiety; it does not say how sharply attention stops at the moiety's edge. Boundary attention coverage restricts the measurement to the boundary region, the moiety's edge atoms together with the one-bond-outside ring, and reports the share of attention in that region falling on the edge atoms themselves. Chance is 0.5 and values above 0.59 indicate a sharp boundary. Median boundary coverage sits at chance at every scale through 1.3B and reaches 0.904 at 1.6B (Table~\ref{tab:bleed}): 95\% of proteins show a sharp moiety boundary at the largest scale, against roughly 1\% at every smaller scale.

\begin{table}[H]
\centering
\caption{Boundary sharpness across model scales. Median boundary attention coverage and per-protein counts in the three bucket categories.}
\label{tab:bleed}
\small
\begin{tabular}{lcccc}
\toprule
Scale & Median boundary coverage & Sharp ($>$0.59) & Soft (0.50--0.59) & Bleeding ($<$0.50) \\
\midrule
42M & 0.501 & 1/90 & 57/90 & 32/90 \\
315M & 0.501 & 1/90 & 66/90 & 23/90 \\
800M & 0.500 & 1/90 & 59/90 & 30/90 \\
1B & 0.501 & 1/90 & 60/90 & 29/90 \\
1.3B & 0.502 & 1/90 & 75/90 & 14/90 \\
\textbf{1.6B} & \textbf{0.904} & \textbf{113/119} & \textbf{1/119} & \textbf{5/119} \\
\bottomrule
\end{tabular}
\end{table}

\section{Computational benchmarks: protocols, datasets, and per-method comparisons}
\label{app:A6}

\subsection{Decontamination}
\label{app:A6-1}
To hold out the evaluation benchmarks (CASP16, FEP+4, OpenFE, MF-PCBA, LIT-PCBA and its subset LIT-PCBA-AF3, COACH420, Davis, and BindingDB), we used MMseqs2~\citep{steinegger_2017_mmseqs2} (\texttt{-{}-min-seq-id 0.9 -c 0.8 -{}-cov-mode 2 -{}-alignment-mode 3 -s 7.5}) to remove from training any protein sharing $\geq$90\% sequence identity over $\geq$80\% coverage with a benchmark target, along with all of its associated protein-ligand pairs. Ligand decontamination used Tanimoto below 0.60 on Morgan fingerprints (radius 2, 2048 bits).

The affinity probe of Appendix~\ref{app:A6-4} is trained on quantitative affinity measurements drawn from PubChem~\citep{kim_2023_pubchem}, ChEMBL~\citep{zdrazil_2024_chembl}, and the BindingDB database~\citep{gilson_2015_bindingdb}, not from the published BindingDB benchmark~\citep{mcnutt_2024_sprint}, which is held out in full. Proteins sharing $\geq$90\% sequence identity with any target in the benchmarks listed above, together with their associated pairs, were removed from the probe's training data before fitting, under the same protocol, so no evaluation target appears in it.

\subsection{Dataset preparation}
\label{app:A6-2}
For MF-PCBA, we selected bioassays and downsampled to 50{,}000 pairs as described in \citet{passaro_2025_boltz}. For LIT-PCBA, we used the AVE set, which removes duplicates and structural bias, a stricter evaluation than the ``full'' split used by some prior work.

\subsection{Affinity benchmark configurations}
\label{app:A6-3}
\begin{figure}[H]
    \centering
    \includegraphics[width=\linewidth]{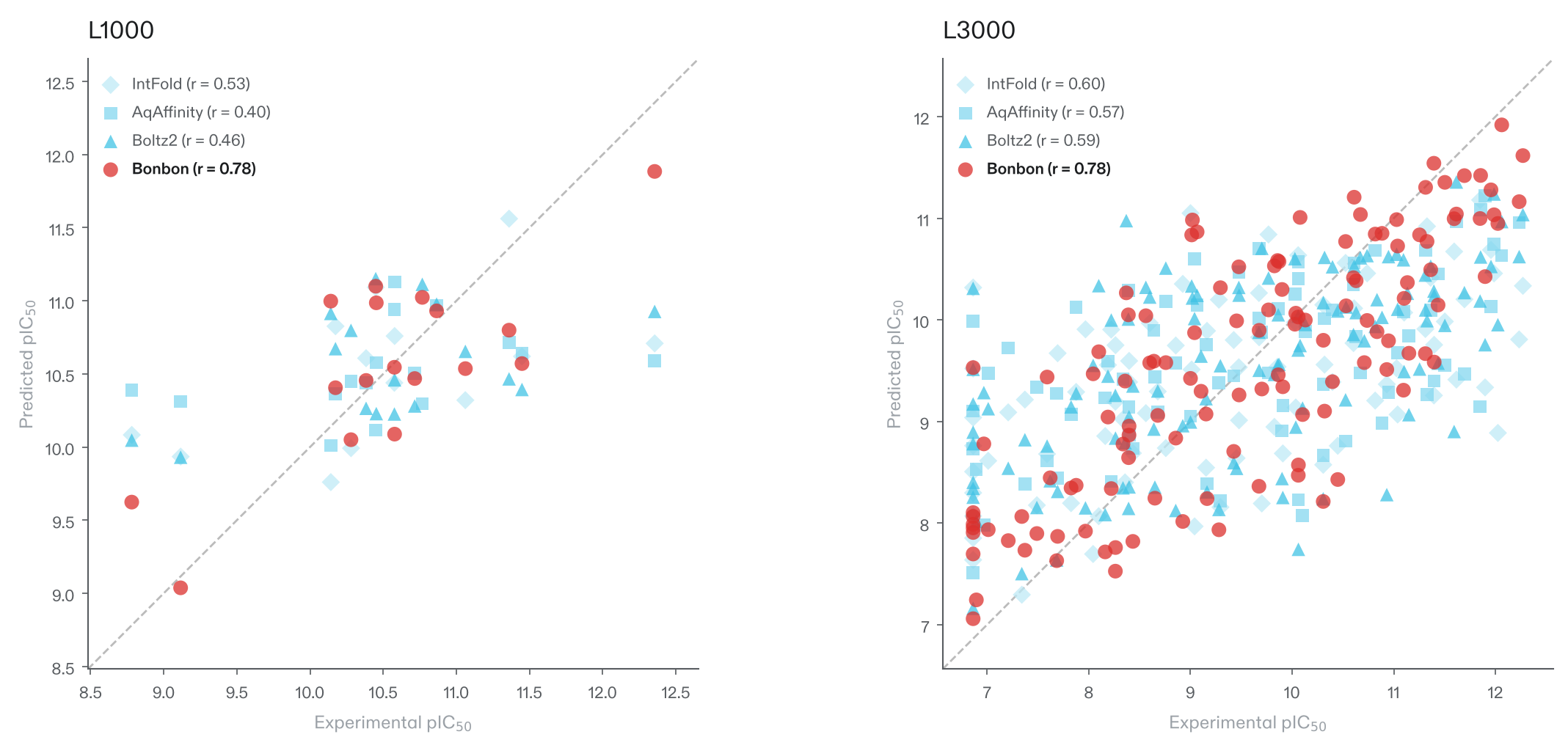}
    \caption{CASP16 binding affinity scatter plots. Predicted vs.\ experimental pIC$_{50}$ on L1000 (left) and L3000 (right). \textsc{Bonbon} ($r = 0.78$ on both) shows tighter clustering along the diagonal than Boltz-2 ($r = 0.46, 0.59$), IntFold ($r = 0.53, 0.60$), and AqAffinity ($r = 0.40, 0.57$). The advantage is consistent across both targets despite differences in binding site topology and ligand chemotype.}
    \label{fig:A3}
\end{figure}

We evaluated on the CASP16 binding affinity challenge comprising 140 protein-ligand pairs across two targets~\citep{gilson_2025_assessment}, the FEP+ four-target subset (CDK2, TYK2, JNK1, p38) as defined in Boltz-2~\citep{passaro_2025_boltz}, and the Open Force Field Protein-Ligand Benchmark~\citep{hahn_2022_best}. FEP+ computes relative binding free energies within a congeneric series; ABFE is the absolute variant and is reported as a separate row in Table~\ref{tab:A14}. All methods in Table~\ref{tab:A14} are compared by Pearson correlation on the same targets, and Pearson $r$ is invariant to scale and offset, so the comparison does not depend on any method being calibrated in absolute terms. We ran Boltz-2 using default settings with predicted structures and its integrated affinity scoring module; all other baselines are reported from published results. We note that our reproduced Boltz-2 CASP16 correlation ($r = 0.52$) differs from the value reported in the original publication ($r = 0.65$). This discrepancy has been independently observed by other groups: IntFold~\citep{intfold_2025}, AqAffinity~\citep{sandboxaq_2026_aqaffinity}, and TerraBind~\citep{terrabind_2026} all report reproduced Boltz-2 correlations in the 0.46--0.59 range on CASP16 subsets. Figure~\ref{fig:A3} plots the CASP16 predictions against experiment for \textsc{Bonbon} and these baselines.
\subsection{Affinity probe}
\label{app:A6-4}
The two-layer MLP probe maps \textsc{Bonbon}'s frozen interaction representations to a scalar affinity value (pKi), the minimal supervised step required to project the representation space onto the affinity axis. No representation fine-tuning is performed. Probe architecture: input dimension = embedding dimension, hidden dimension = 512, output dimension = 1. Training used AdamW with default hyperparameters and a Huber objective for 3{,}800 steps ($\sim$15 minutes) on the combined PubChem, ChEMBL, and BindingDB measurements of Appendix~\ref{app:A6-1}, with no early stopping.

\subsection{Binary binding prediction: full per-method comparison}
\label{app:A6-5}
\begin{table}[H]
\centering
\caption{Binary binding prediction. Bold indicates best performance. ``---'' indicates not reported. $\dagger$ = structure-informed (uses protein structure embeddings). $\ddagger$ = structure-based (uses full 3D coordinates or docking). * AVE split (deduplicated). ** SPRINT evaluated on ``full'' split without deduplication~\citep{sprint_2024}.}
\label{tab:A12}
\small
\begin{tabular}{lcccccccc}
\toprule
Benchmark & Metric & \textsc{Bonbon} & Boltz-2$^{\ddagger}$ & IntFold$^{\ddagger}$ & SPRINT$^{\dagger}$ & DrugCLIP$^{\dagger}$ & Docking$^{\ddagger}$ \\
\midrule
Davis & AUPR & \textbf{0.825} & 0.46 & 0.58 & 0.51 & --- & --- \\
BindingDB & AUPR & \textbf{0.752} & 0.53 & 0.67 & 0.72 & --- & --- \\
MF-PCBA & AUPR & \textbf{0.081} & 0.025 & --- & --- & --- & 0.016 \\
LIT-PCBA* & EF1\% & \textbf{11.4} & --- & --- & 10.8** & 5.5 & 3.4--4.6 \\
\bottomrule
\end{tabular}
\end{table}

\begin{table}[H]
\centering
\caption{Enrichment factor at 1\% on LIT-PCBA-AF3~\citep{shen_2025_unlocking}, a five-target subset of LIT-PCBA restricted to libraries of fewer than 100{,}000 compounds.}
\label{tab:litpcba-af3}
\small
\begin{tabular}{lc}
\toprule
Method & EF1\% \\
\midrule
\textbf{\textsc{Bonbon}} & \textbf{16.6} \\
Boltz-2 & 9.3 \\
Protenix & 7.6 \\
Glide & 6.8 \\
ConPlex & 4.3 \\
Gnina & 2.8 \\
SPRINT & 1.8 \\
\bottomrule
\end{tabular}
\end{table}

\subsection{Binding affinity prediction: per-target results and full per-method comparison}
\label{app:A6-6}

\begin{table}[H]
\centering
\caption{Binding affinity prediction, full comparison (Pearson correlation). Methods grouped by type. Bold indicates the best value in each column. $\dagger$ = independently reproduced. ``---'' indicates not reported.}
\label{tab:A14}
\small
\begin{tabular}{llccc}
\toprule
Method & Type & CASP16 & FEP+4 & OpenFE \\
\midrule
FEP+ & Physics ($>$1h) & --- & 0.78 & 0.72 \\
ABFE & Physics ($>$1h) & --- & 0.75 & 0.73 \\
OpenFE-calc & Physics ($>$1h) & --- & 0.66 & 0.62 \\
FMO & Physics ($<$1h) & --- & 0.55 & --- \\
MM/PBSA & Physics ($<$1h) & --- & 0.18 & --- \\
Chemgauss4 & Docking & --- & 0.26 & --- \\
Ruby Max~\citep{genesis_2026_ruby} & Structure-based ML & 0.77 & 0.86 & 0.74 \\
IsoDDE~\citep{isomorphic_2026} & Structure-based ML & 0.75 & 0.85 & 0.73 \\
Nesso-1~\citep{valence_2026_nesso} & Structure-based ML & 0.65 & 0.81 & 0.66 \\
Boltz-2~\citep{passaro_2025_boltz} & Structure-based ML & $0.52^{\dagger}$ & 0.66 & 0.62 \\
IntFold~\citep{intfold_2025} & Structure-based ML & 0.54 & --- & --- \\
AqAffinity~\citep{sandboxaq_2026_aqaffinity} & Structure-based ML & 0.49 & 0.50 & 0.33 \\
GAT & Structure-based ML & --- & 0.40 & --- \\
BACPI & Structure-based ML & 0.38 & 0.14 & 0.28 \\
Haddock & Docking & 0.38 & --- & --- \\
Haiping & Sequence-based ML & 0.54 & --- & --- \\
\textbf{\textsc{Bonbon}} & \textbf{Sequence-based ML} & \textbf{0.78} & \textbf{0.88} & \textbf{0.76} \\
\bottomrule
\end{tabular}
\end{table}

\subsection{Inference throughput}
\label{app:A6-7}
\begin{figure}[H]
\centering
\includegraphics[width=0.75\linewidth]{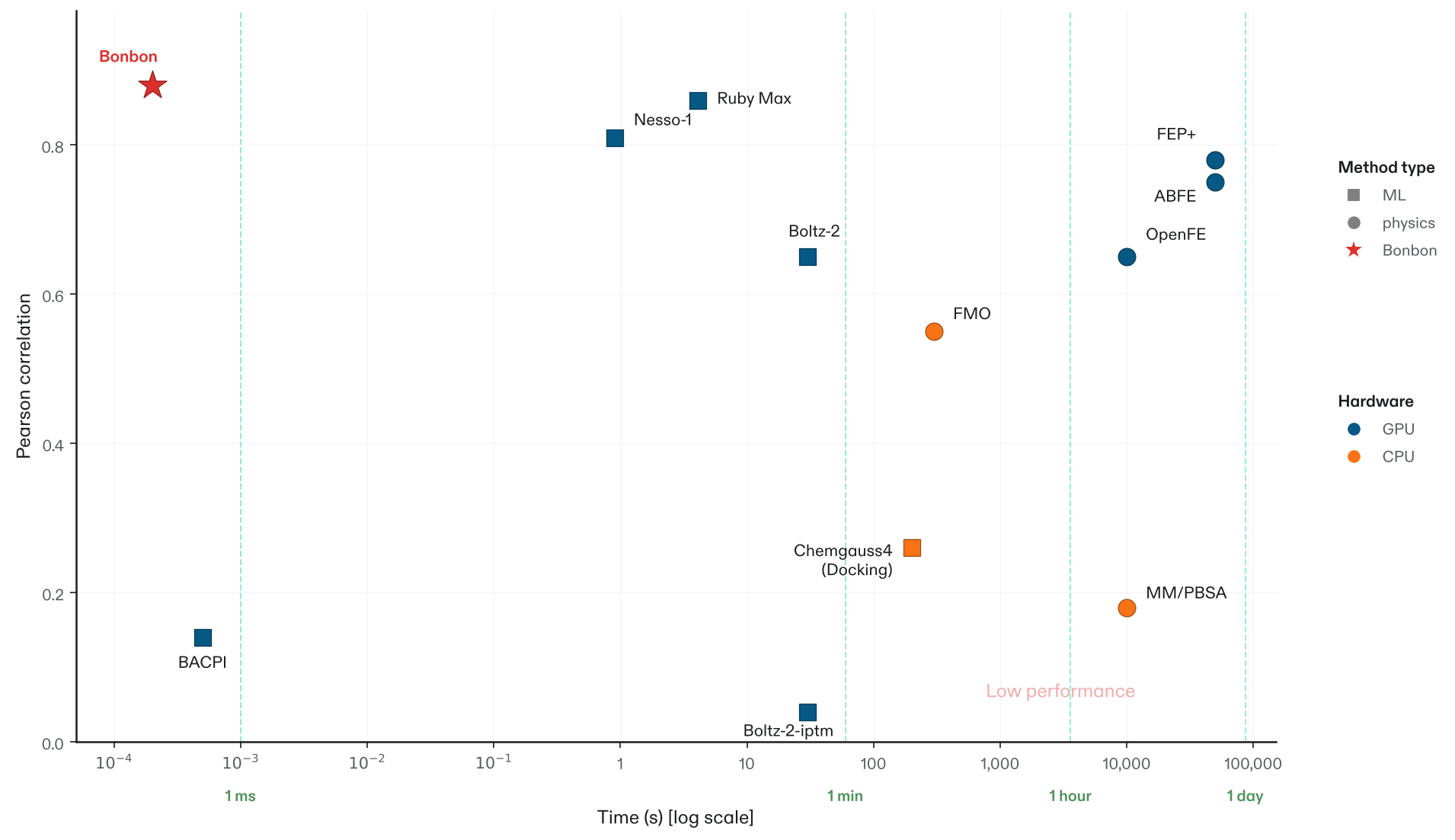}
\caption{Accuracy versus inference time on FEP+4: \textsc{Bonbon} achieves $r = 0.88$ in 0.08 seconds, against seconds for structure-dependent machine learning methods and hours for physics-based methods. \textsc{Bonbon} is both the most accurate and the fastest method evaluated.}
\label{fig:speed-accuracy}
\end{figure}

Table~\ref{tab:A15} reports measured and extrapolated throughput; Figure~\ref{fig:speed-accuracy} places \textsc{Bonbon} against structure-dependent and physics-based methods on the speed-accuracy plane.

\begin{table}[H]
\centering
\caption{Inference throughput on a single H100 GPU. Boltz-2 was empirically measured at 20.6\,s/pair on the same hardware, consistent with the 20\,s/pair reported in the Boltz-2 paper on FEP+ using an H100 GPU. * = extrapolated from measured per-pair rate. ** = extrapolated for both models.}
\label{tab:A15}
\small
\begin{tabular}{lllccc}
\toprule
Task & Pairs & Mode & \textsc{Bonbon} & Boltz-2 & Speedup \\
\midrule
FEP+4 & 87 & On-the-fly & 0.08\,s & 3{,}867\,s & ${\sim}48{,}600\times$ \\
MF-PCBA (10 targets) & 500K & On-the-fly & 5\,min & ${\sim}4$\,months* & ${\sim}34{,}700\times$ \\
LIT-PCBA & 2.65M & On-the-fly & 18\,min & ${\sim}1.75$\,years* & ${\sim}51{,}100\times$ \\
Proteome $\times$ DrugBank & 1.66B & Cached & 11\,hours** & ${\sim}1{,}100$\,GPU-years** & ${\sim}873{,}000\times$ \\
\bottomrule
\end{tabular}
\end{table}

\section{Prospective experimental validation: methods and per-target detail}
\label{app:A7}

\paragraph{Attribution and reproducibility statement.} All biochemical assays in this section were performed at an external contract research organization (CRO) under protocols specified in advance by the authors. Compounds were selected and ranked by the authors using \textsc{Bonbon} scores (\ref{app:A7-1}) and delivered to the CRO in a randomized format; the CRO returned raw dose-response data, which the authors then mapped back to compound identities for hit analysis. The CRO did not filter compounds, repeat assays, or alter protocols. The CRO's identity is withheld under a confidentiality agreement; per-target protocols (\ref{app:A7-5}--\ref{app:A7-13}) reference standard assay formats and published methods and are sufficient to reproduce all results at any comparably equipped facility.

\subsection{Compound ranking and selection}
\label{app:A7-1}
For each target, we used \textsc{Bonbon} to rank molecules produced by a separate generative method under synthetic-accessibility constraints; that method is outside the scope of this paper and is the subject of a separate technical report. \textsc{Bonbon}'s role in this experiment is ranking alone. Compounds with Tanimoto similarity $\geq 0.60$ to any compound in \textsc{Bonbon}'s training data were removed (Morgan fingerprints, radius 2, 2048 bits). PAINS-pattern compounds~\citep{baell_2010_new} were filtered. The remaining top-128 compounds by \textsc{Bonbon} score were clustered using hierarchical agglomerative clustering (distance = $1 - $ Tanimoto similarity, fusion cutoff = 0.5). Twenty-four compounds were selected from the clustered pool to maximize chemical diversity within the top-ranked set. No additional human selection or medicinal chemistry triage was applied. For MMP-9 one compound and for CDK1 two compounds could not be obtained, yielding 23 and 22 tested compounds respectively and 213 compounds across the nine targets. All delivered compounds were profiled in a single-dose primary screen, followed by ten-point dose-response confirmation for IC$_{50}$ determination (\ref{app:A7-2}). Confirmed inhibitors are compounds active in both the primary and dose-response assays. Dose-response hit rates are reported over all delivered compounds.

\subsection{Compound sourcing and handling}
\label{app:A7-2}
Compounds were synthesized by a contract research organization and delivered as 10\,mM stocks in DMSO, verified to at least 90\% purity by LC-MS. Stocks were stored at $-80\,^{\circ}$C (long-term) or $-20\,^{\circ}$C (working) in desiccated storage and thawed at room temperature before use. All compounds were first assayed in a single-dose primary screen in duplicate at 10\,$\mu$M and 1\% DMSO. Compounds meeting the progression criterion were assayed in dose-response in duplicate across a ten-point, three-fold serial dilution from a top final concentration of 100\,$\mu$M to 5\,nM, with reference inhibitors titrated in parallel on every plate. Per-concentration responses were normalized to percent activity relative to DMSO vehicle controls after subtraction of no-enzyme background; for the protease assays the response was the initial reaction velocity normalized to vehicle. Normalized responses were fit against compound concentration with a four-parameter logistic (sigmoidal, variable-slope) model,
\begin{equation*}
Y = \mathrm{Bottom} + \frac{\mathrm{Top} - \mathrm{Bottom}}{1 + 10^{(\log \mathrm{IC}_{50} - \log[\mathrm{compound}]) \cdot \mathrm{HillSlope}}},
\end{equation*}
reporting the midpoint potency IC$_{50}$ (and pIC$_{50} = -\log_{10}$ IC$_{50}$ [M]) together with the Hill slope. A compound counts as a confirmed inhibitor if its dose-response series returns a determinable IC$_{50}$ from this fit; no compound that progressed to dose-response was excluded from the analysis.

\subsection{Novelty analysis}
\label{app:A7-3}
Known actives for each target were aggregated from ChEMBL and BindingDB. A hit counts as a novel molecule only if it clears three criteria: it is not a duplicate of any known active; its maximum ECFP4 Tanimoto (Morgan, radius 2, 2048 bits) to any known active is below 0.45; and it is not a single-substituent analog of any known active by matched-molecular-pair analysis~\citep{hussain_rea_2010}, with the changing fragment limited to at most 10 heavy atoms and the constant core required to be at least twice its size, following standard mmpdb limits~\citep{dalke_2018}.

\subsection{Per-target results}
\label{app:A7-4}
\begin{figure}[H]
\centering
\includegraphics[width=\linewidth]{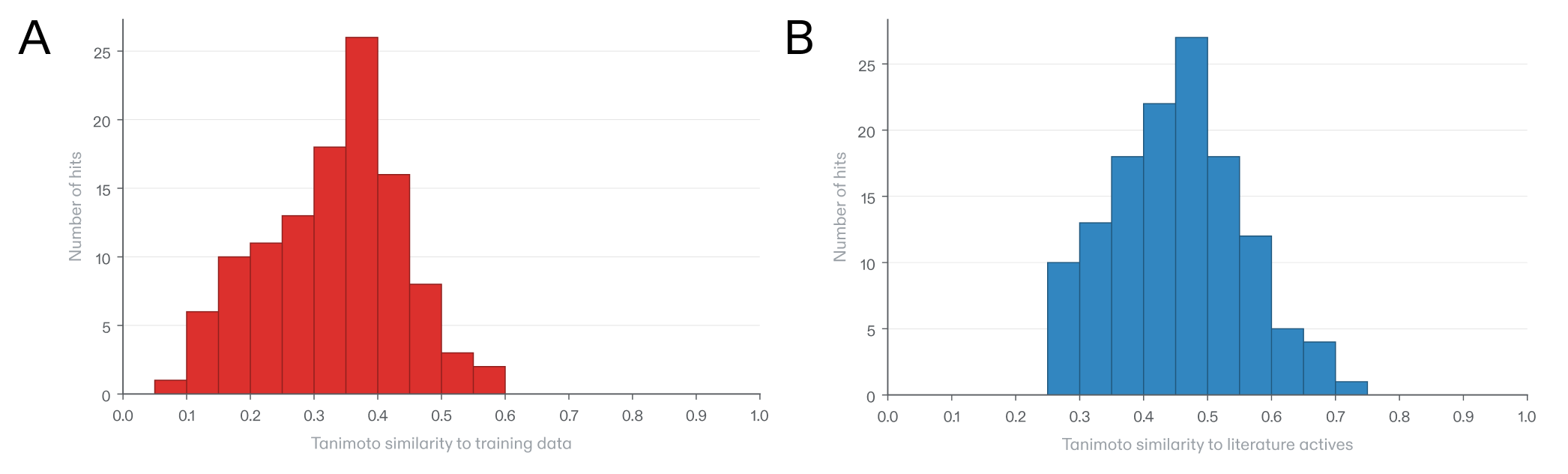}
\caption{\textbf{(a)} Tanimoto similarity of the confirmed inhibitors to the nearest molecule in the training data; all fall below 0.60. KDR/VEGFR2 and ELANE are excluded: neither protein appears in the training corpus, nor does any protein above 90\% sequence identity to either, so neither target has training molecules to compare against. \textbf{(b)} Tanimoto similarity of the confirmed inhibitors to the nearest known active for their target, across all nine targets.}
\label{fig:tanimoto}
\end{figure}

Table~\ref{tab:A7-targets} reports, for each of the nine targets, the number of compounds tested and confirmed, the IC$_{50}$ range, and the Tanimoto range to the nearest known active. Figure~\ref{fig:tanimoto} shows the corresponding per-compound distributions against training data and against known actives.

\begin{table}[H]
\centering\small
\caption{Per-target dose-response results across the nine-target panel. Confirmed inhibitors are compounds active in both the primary and dose-response assays. Tanimoto is the range of maximum ECFP4 (Morgan, radius 2, 2048 bits) similarity of each confirmed inhibitor to any known active for that target.}
\label{tab:A7-targets}
\begin{tabular}{lrrll}
\toprule
Target & Tested & Confirmed & IC$_{50}$ range & Tanimoto \\
\midrule
TYK2        & 24 & 23 & 146\,nM--40\,$\mu$M    & 0.342--0.567 \\
PRMT5       & 24 & 16 & 186\,nM--143\,$\mu$M   & 0.479--0.597 \\
JAK1        & 24 & 21 & 3\,nM--28\,$\mu$M      & 0.385--0.684 \\
CDK1        & 22 & 22 & 114\,nM--23\,$\mu$M    & 0.279--0.706 \\
HDAC6       & 24 & 15 & 115\,nM--22\,$\mu$M    & 0.278--0.625 \\
MNK2        & 24 & 19 & 39\,nM--38\,$\mu$M     & 0.269--0.492 \\
MMP-9       & 23 &  3 & 480\,nM--1.3\,$\mu$M   & 0.443--0.563 \\
KDR/VEGFR2  & 24 & 10 & 38\,nM--82\,$\mu$M     & 0.429--0.617 \\
ELANE       & 24 &  2 & 1.22--29.4\,$\mu$M     & 0.407--0.500 \\
\midrule
Overall     & 213 & 131 & & \\
\bottomrule
\end{tabular}
\end{table}

\subsection{TYK2 kinase assay}
\label{app:A7-5}
TYK2 activity was measured in a radiometric filter-binding assay~\citep{anastassiadis_2011_comprehensive}. TYK2 transfers the $\gamma$-phosphate from a mixture of ATP and [$\gamma$-${}^{33}$P]-ATP to the peptide substrate KKSRGDYMTMQIG (20\,$\mu$M). Reactions were run in standard kinase buffer (20\,mM HEPES pH 7.5, 10\,mM MgCl$_2$, 1\,mM EGTA, 0.01\% Brij-35, 0.02\,mg/mL BSA, 0.1\,mM Na$_3$VO$_4$, 2\,mM DTT, 1\% DMSO) with ATP at its apparent $K_m$ (10\,$\mu$M); compounds were delivered by acoustic transfer (Echo 550) and pre-incubated 20 minutes, the reaction then proceeded 2 hours at 25$^{\circ}$C, and ${}^{33}$P-substrate was captured on P81 ion-exchange filter paper and washed in 0.75\% phosphoric acid. Staurosporine was the reference inhibitor (IC$_{50}$ = 0.39\,nM).

\subsection{PRMT5 methyltransferase assay}
\label{app:A7-6}
PRMT5/MEP50 activity was measured in a radiometric filter-binding assay. Transfer of a tritiated methyl group from [${}^{3}$H]-S-adenosylmethionine (SAM, 1\,$\mu$M) to histone H2A (5\,$\mu$M) proceeded for 60 minutes at 30$^{\circ}$C in 50\,mM Tris-HCl (pH 8.5), 5\,mM MgCl$_2$, 50\,mM NaCl, 1\,mM DTT, 0.01\% Brij-35, and 1\% DMSO. Compounds were delivered by acoustic transfer (Echo 550) and pre-incubated 20 minutes. Reactions were quenched and transferred to nylon filter paper, which was washed in PBS to remove unreacted radioactive SAM. S-adenosyl-L-homocysteine (SAH) was the reference inhibitor (IC$_{50}$ = 3.4\,$\mu$M).

\subsection{JAK1 kinase assay}
\label{app:A7-7}
JAK1 activity was measured in a radiometric filter-binding assay. JAK1 transfers the $\gamma$-phosphate from a mixture of ATP and [$\gamma$-${}^{33}$P]-ATP to a poly[Glu:Tyr] (4:1) substrate (0.2\,mg/mL). Reactions were run in standard kinase buffer with ATP at its apparent $K_m$ (15\,$\mu$M); compounds were delivered by acoustic transfer (Echo 550) and pre-incubated 20 minutes, the reaction then proceeded 2 hours at 25$^{\circ}$C, and ${}^{33}$P-substrate was captured on P81 ion-exchange filter paper and washed in 0.75\% phosphoric acid. Staurosporine was the reference inhibitor (IC$_{50}$ = 0.76\,nM).

\subsection{CDK1 kinase assay}
\label{app:A7-8}
CDK1 activity was measured in a radiometric filter-binding assay of the CDK1/cyclin B complex. CDK1 transfers the $\gamma$-phosphate from a mixture of ATP and [$\gamma$-${}^{33}$P]-ATP to retinoblastoma protein substrate (5\,$\mu$M). Reactions were run in standard kinase buffer with ATP at its apparent $K_m$ (50\,$\mu$M); compounds were delivered by acoustic transfer (Echo 550) and pre-incubated 20 minutes, the reaction then proceeded 2 hours at 25$^{\circ}$C, and ${}^{33}$P-substrate was captured on P81 ion-exchange filter paper and washed in 0.75\% phosphoric acid. Staurosporine was the reference inhibitor (IC$_{50}$ = 1.8\,nM).

\subsection{HDAC6 deacetylase assay}
\label{app:A7-9}
HDAC6 deacetylase activity was measured in a two-step fluorogenic assay using the acetylated peptide substrate from p53 residues 379--382 (RHKK(Ac)-AMC, 10\,$\mu$M); deacetylation followed by trypsin digestion frees 7-amino-4-methylcoumarin (AMC) for quantification. Reactions were run in 50\,mM Tris (pH 8.0), 137\,mM NaCl, 2.7\,mM KCl, 1\,mM MgCl$_2$, 1\,mg/mL BSA, and 1\% DMSO. Compounds were delivered by acoustic transfer (Echo 550) and pre-incubated 20 minutes; the reaction then proceeded 1 hour at 30$^{\circ}$C. Trichostatin A was the reference inhibitor (IC$_{50}$ = 1.35\,nM).

\subsection{MNK2 kinase assay}
\label{app:A7-10}
MNK2 activity was measured in a radiometric filter-binding assay. MNK2 transfers the $\gamma$-phosphate from a mixture of ATP and [$\gamma$-${}^{33}$P]-ATP to myelin basic protein from porcine brain (20\,$\mu$M). Reactions were run in standard kinase buffer with ATP at its apparent $K_m$ (100\,$\mu$M); compounds were delivered by acoustic transfer (Echo 550) and pre-incubated 20 minutes, the reaction then proceeded 2 hours at 25$^{\circ}$C, and ${}^{33}$P-substrate was captured on P81 ion-exchange filter paper and washed in 0.75\% phosphoric acid. Staurosporine was the reference inhibitor (IC$_{50}$ = 18.6\,nM)~\citep{jin_2021_targeting}.

\subsection{MMP-9 protease assay}
\label{app:A7-11}
MMP-9 activity was measured in a continuous fluorogenic protease assay using the MMP-9 catalytic domain (residues 88--438) and the FRET peptide substrate 5-FAM/QXL (5\,$\mu$M). Compounds were delivered by acoustic transfer (Echo 550) and pre-incubated 20 minutes; the reaction was started by substrate addition and monitored as a time course, with the initial linear portion of the slope (signal/min) analyzed. GM6001 (ilomastat) was the reference inhibitor (IC$_{50}$ = 0.13\,nM).

\subsection{ELANE (neutrophil elastase) protease assay}
\label{app:A7-12}
ELANE activity was measured in a continuous fluorogenic protease assay using the peptide substrate MeOSuc-AAPV-AMC (10\,$\mu$M). Compounds were delivered by acoustic transfer (Echo 550) and pre-incubated 20 minutes; the reaction was started by substrate addition and monitored as a time course, with the initial linear portion of the slope (signal/min) analyzed. Sivelestat was the reference inhibitor (IC$_{50}$ = 3.4\,nM)~\citep{voynow_2021_neutrophil}.

\subsection{KDR/VEGFR2 kinase assay}
\label{app:A7-13}
KDR/VEGFR2 activity was measured in a radiometric filter-binding assay. KDR/VEGFR2 transfers the $\gamma$-phosphate from a mixture of ATP and [$\gamma$-${}^{33}$P]-ATP to a poly[Glu:Tyr] (4:1) substrate (0.2\,mg/mL). Reactions were run in standard kinase buffer with ATP at its apparent $K_m$ (2.5\,$\mu$M); compounds were delivered by acoustic transfer (Echo 550) and pre-incubated 20 minutes, the reaction then proceeded 2 hours at 25$^{\circ}$C, and ${}^{33}$P-substrate was captured on P81 ion-exchange filter paper and washed in 0.75\% phosphoric acid. Staurosporine was the reference inhibitor (IC$_{50}$ = 5.4\,nM).

\end{document}